\documentclass[%
 reprint,
groupedaddress,
bibnotes,
 amsmath,
 amssymb,
 aps,
 prl,
]{revtex4-2}

\newcommand{\networkName}[0]{MIAE}
\newcommand{\methodName}[0]{MIDAS}

\usepackage{mathtools}
\usepackage{bm} % bold math
\usepackage{graphicx}
\usepackage{orcidlink}
\usepackage{soul}

\usepackage{xcolor}
\hypersetup{
    colorlinks,
    linkcolor={blue},
    citecolor={blue},
    urlcolor={blue}
}

\usepackage{etoolbox}
\patchcmd{\section}
  {\centering}
  {\raggedright}
  {}
  {}

\begin{document}
% \linenumbers
\title{Mechanics-Informed Autoencoder Enables Automated Detection and Localization of Unforeseen Structural Damage}
%%%%%%%%%%%%%%%%%%%%%%%%%%%%%%%%%%%%%%%%%%%%%%%%%%%%%%%%%%%%%%%%%%%%%%%%
\author{Xuyang Li\orcidlink{0000-0002-6846-0906}}
\author{Hamed Bolandi\orcidlink{0000-0002-1664-3663}}
\author{Mahdi Masmoudi\orcidlink{0009-0003-3326-9305}}
\author{Talal Salem\orcidlink{0000-0002-4983-4431}}
\author{Ankush Jha\orcidlink{0009-0005-4376-4056}}
\author{Nizar Lajnef\orcidlink{0000-0001-9578-1054}}
\author{Vishnu Naresh Boddeti\orcidlink{0000-0002-8918-9385}}
\affiliation{Michigan State University, East Lansing, MI 48824, USA}
%%%%%%%%%%%%%%%%%%%%%%%%%%%%%%%%%%%%%%%%%%%%%%%%%%%%%%%%%%%%%%%%%%%%%%%%
\date{\today}

\begin{abstract}
Structural health monitoring (SHM) ensures the safety and longevity of structures like buildings and bridges. As the volume and scale of structures and the impact of their failure continue to grow, there is a dire need for SHM techniques that are scalable, inexpensive, can operate passively without human intervention, and are customized for each mechanical structure without the need for complex baseline models. We present \methodName, a novel ``deploy-and-forget" approach for automated detection and localization of damage in structures. It is a synergistic integration of entirely passive measurements from inexpensive sensors, data compression, and a mechanics-informed autoencoder. Once deployed, \methodName\ continuously learns and adapts a bespoke baseline model for each structure, learning from its undamaged state's response characteristics. After learning from just 3 hours of data, it can autonomously detect and localize different types of unforeseen damage. Results from numerical simulations and experiments indicate that incorporating the mechanical characteristics into the autoencoder allows for up to a 35\% improvement in the detection and localization of minor damage over a standard autoencoder. Our approach holds significant promise for reducing human intervention and inspection costs while enabling proactive and preventive maintenance strategies. This will extend the lifespan, reliability, and sustainability of civil infrastructures.
\end{abstract}

\maketitle
\begin{figure*}[!ht]
\centering
\includegraphics[width=\textwidth]{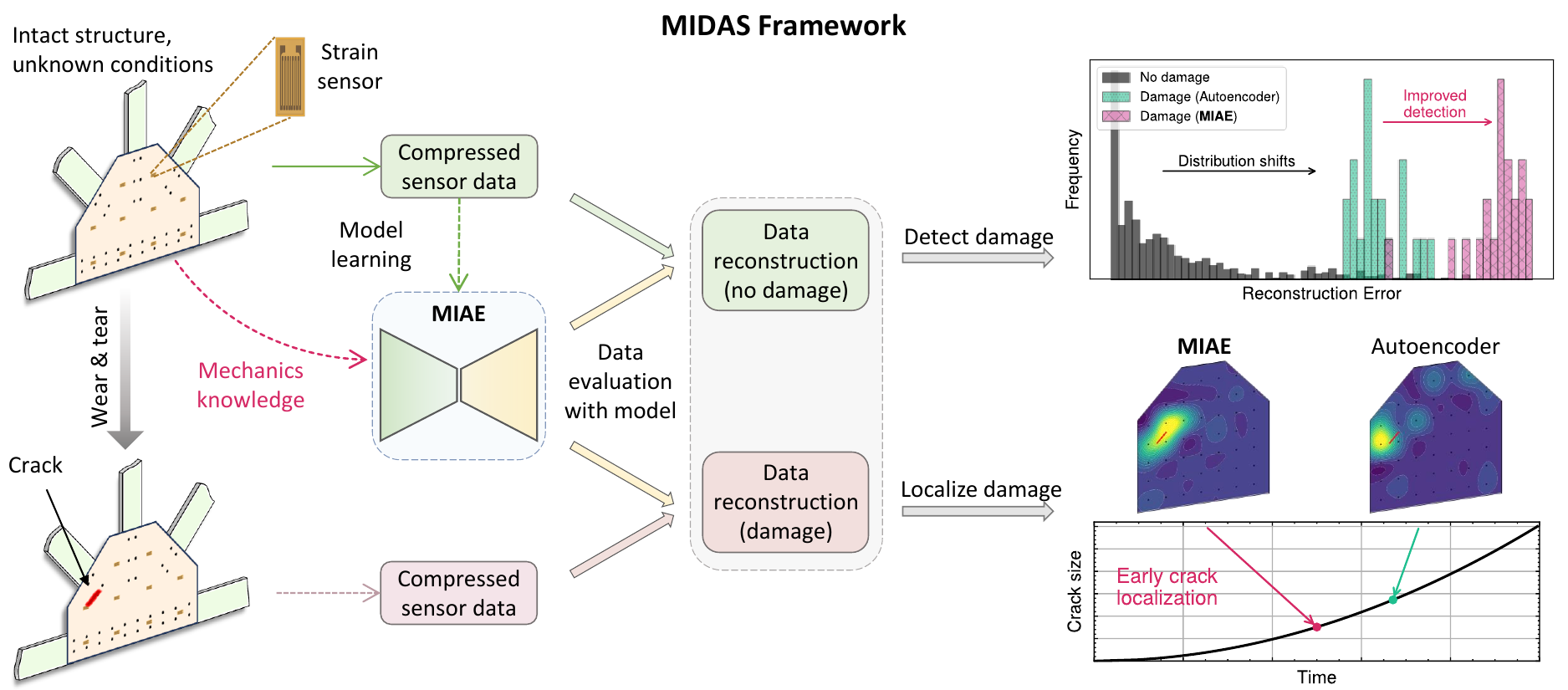}
\caption{\textbf{Overview of \methodName{}, the automated structural damage detection and localization framework.} Raw structural response data from the sensors are compressed, and \networkName\ is trained purely on the response from the structure's undamaged state. No additional information is leveraged besides the pairwise mechanical relations between the strain responses. Once trained, the distribution of reconstruction errors between the network's input and output on the training data serves as a reference representation of an intact structure's response. After deployment, the trained model processes data from the sensors, and resultant reconstruction errors are compared to the reference error distribution to detect and localize potential damage. An observable shift in reconstruction errors (top right) highlights the detection of damage. The incorporated mechanical knowledge notably amplifies the distribution shift, significantly enhancing damage detection at an early stage. Sensor-wise error comparisons are interpolated (heatmaps at the bottom right) to localize anomalies representing the onset of damage\label{fig:1}}
\end{figure*}

Structural health monitoring plays a vital role in monitoring and ensuring the safety and reliability of various engineering systems. Poor monitoring and maintenance can lead to severe damage or even catastrophic failures of structures. Numerous structural failures have occurred despite frequent manual inspections and the adoption of many active sensing technologies over the years. For instance, a severe crack in the I-40 Bridge in Memphis went undetected for years before being discovered in 2021~\cite{pezeshk2021data}, resulting in long-term road closure, substantial economic losses, and significant safety concerns among the public. Similarly, in 2022, a bridge in Pittsburgh collapsed due to the corrosion and deterioration of the bridge legs~\cite{carnahan2022pittsburgh}, damaging several vehicles and causing many injuries. Preventing such incidents as the built environment scales and ages necessitates the development of passive, inexpensive, and continuous structural monitoring techniques, with the ultimate aim of detecting, localizing, and identifying different types of damage at an early stage. Such solutions would complement existing active and costly manual inspections.

\textcolor{black}{SHM systems often employ sensors to measure physical quantities such as strain, vibration, and temperature. The measurements are coupled with a numerical model to infer the structure's health condition. Real-world deployment of SHM has to contend with multiple challenges due to the complexity and diversity of structures, sensors, and damage scenarios. First, detecting and localizing damages as early as possible is critical to extend the structure's longevity. However, minor damage, hidden or distributed in the structure, may not readily manifest in the sensor data and cannot be identified by the numerical model. Second, due to the sheer diversity of structures and associated damage they may endure, SHM methods have to contend with unknown or novel damage without being able to rely on prior knowledge or annotated data. Third, multiple sensors are typically used at different locations on the structure. Seamless SHM will require a combination of inexpensive passive sensors and algorithms that can simultaneously and effectively utilize data from multiple sensors.}

\textcolor{black}{While many solutions have been developed for SHM, existing solutions are limited in multiple respects. They either need active measurements~\cite{capineri2021ultrasonic, tang2021guided, lissenden2015use, chandarana2017early, song2022improved}, detect but do not localize damage~\cite{chandarana2017early, lissenden2015use, daneshvar2021early, bakhary2010substructuring, betti2015damage, li2023real, hasni2017new, esfandiari2020structural}, or employ technology that is accurate but very complex and expensive, such as guided waves~\cite{sawant2023unsupervised, capineri2021ultrasonic, tang2021guided} and acoustic emissions~\cite{chandarana2017early, song2022improved}. Furthermore, some are based on predefined damage features or thresholds~\cite{sohn2003review, khan2016integration}, designed to model data from a single sensor or do not take the domain attributes of the structure and the sensor placement into account when detecting or localizing damage~\cite{wang2021unsupervised, jiang2021decentralized}, or are limited to identify known types of damage~\cite{abdeljaber2017real,ma2022real} only.}

\textcolor{black}{In the broader context of structural engineering, machine learning methods are increasingly being relied upon for addressing many problems. For instance, Physics-Informed Neural Networks (PINNs)~\cite{raissi2018hidden, raissi2019physics, bolandi2023physics}, which leverage both data and knowledge of the underlying physics, and Graph Neural Networks (GNNs)~\cite{parisi2024use, song2023elastic, chou2024structgnn, bloemheuvel2021computational, zhan2023novel} are commonly being employed for forward and inverse problems. These solutions promise significant computational gains over traditional numerical methods. However, the need for precise knowledge of the governing equations, parameters, loading, etc., limits their applicability for detecting and localizing damage in the real world, where such information is usually unavailable.}

Current SHM solutions instead rely on more traditional machine learning (ML) such as support vector machines~\cite{hasni2017detection} for steel bridge structures, neural networks for buildings~\cite{gonzalez2008seismic}, concrete slabs~\cite{bakhary2010substructuring}, pavement~\cite{ma2022real}, and steel frames~\cite{betti2015damage}, and recurrent neural networks \cite{rautela2021ultrasonic}, long short-term memory (LSTM) and gated recurrent units~\cite{choe2021sequence} to detect, localize, and quantify structural defects. Such solutions have also been proposed to detect damage in gusset plates~\cite{xu2019multistage, gulgec2019convolutional, mustafa2023evaluation, li2022methods}, bridges~\cite{wu2018damage, li2021deep, heo2018experimental}, highway sections~\cite{chen2019damage}, and railways~\cite{azim2021data, azim2021development}. An extended discussion of machine learning-based approaches for SHM can be found in Supplementary note 4.

\textcolor{black}{The primary drawback of the aforementioned body of work is their need for annotated sensor data with labels corresponding to normal or damaged operating conditions. Obtaining such annotations in large quantities and for each deployment is costly and impractical. Furthermore, models learned through explicit supervision often fail to generalize to unseen damage scenarios. A few unsupervised anomaly detection approaches have also been developed with a focus on autoencoders~\cite{rastin2021unsupervised, ni2020deep, giglioni2023autoencoders} and principal component analysis (PCA)~\cite{khoshnoudian2017structural, cao2019baseline, esfandiari2020structural, sen2019effectiveness}. Besides detection, a limited number of approaches focused on damage localization with finite element models~\cite{zhang2021structural}, convolutional neural networks~\cite{zhang2021damage, cofre2019deep}, and autoencoders~\cite{jiang2021decentralized}. These existing unsupervised methods~\cite{wang2021unsupervised, jiang2021decentralized}, however, are typically designed to model data from a single sensor or do not take the domain attributes of the structure and the sensor placement into account when detecting or localizing damage.} 

We propose Mechanics-Informed Damage Assessment of Structures (\methodName), a near-real-time SHM framework for automated damage detection and localization. Our solution is based on the premise that sensor data collected from a structure during its regular operation represents its expected behavior, and any deviation from this behavior indicates potential damage. A structure we wish to assess for damage is instrumented with sensors, and data from its undamaged state is collected to establish the reference (baseline) for damage detection through unsupervised learning. The established reference can be employed to detect and localize damage. \textcolor{black}{This solution affords adaptation to known and unknown damage across diverse structures like gusset plates and beam columns.}

\textcolor{black}{The key contribution of \methodName\ is the seamless integration of inexpensive sensors, data pre-processing in the form of compression, and a customized autoencoder called Mechanics-Informed Autoencoder (MIAE). From a sensor perspective, our solution is agnostic to the sensor technology and can even employ wireless sensors~\cite{alavi2016intelligent, alavi2016damage, hasni2017new, hasni2017continuous}, which are becoming cost-effective and widely used today. These sensors are easier to install and maintain and are often self-powered, rendering them very effective for long-term monitoring. From a pre-processing perspective, we leverage the on-device data compression (edge computing)~\cite{salehi2021comprehensive, alavi2016damage, hasni2017continuous} offered by modern sensors and use a highly (temporally) compressed version of the raw sensor data. Subsequently, variations due to environmental or loading fluctuations are filtered away by the compression. Therefore, any abnormal patterns in the data are indicative of damage. From the neural network perspective, we adopt an autoencoder that learns a compact representation of the data streams from multiple sensors while incorporating the mechanical relations between their strain responses. Such a design significantly enhances the detection and localization of damage in the structure.}

Figure \ref{fig:1} shows an overview of \methodName{}. Damage detection is achieved by comparing the reconstruction error of the instantaneous sensor data in time windows with that of the undamaged baseline. To localize the damage, we further compute the norms of reconstruction errors at each sensor and interpolate them between the sensors. This approach does not require data from damaged structures for training, which is a significant advantage of our method, given that collecting realistic damaged data on large-scale structures is practically infeasible. Other techniques that use simulated damage scenarios are often inaccurate and impractical for real-time applications due to the constant need for re-calibration. In contrast, \methodName\ relies solely on reference data to establish an intact model reference and detect damage by tracking deviations from this reference. Furthermore, with the integrated mechanical knowledge, \networkName\ significantly improves its performance in detecting and localizing damage early when it is minor.

\section{Results}
We evaluate the effectiveness of \methodName\ in three ways: (i) numerical simulation of a gusset plate, (ii) experimental validation on a gusset plate, and (iii) experimental validation on a beam-column structure. Beyond these structures, \methodName\ can be readily employed to monitor the health of other kinds of structural components.

\begin{figure*}[!ht]
\centering
\includegraphics[width=\textwidth]{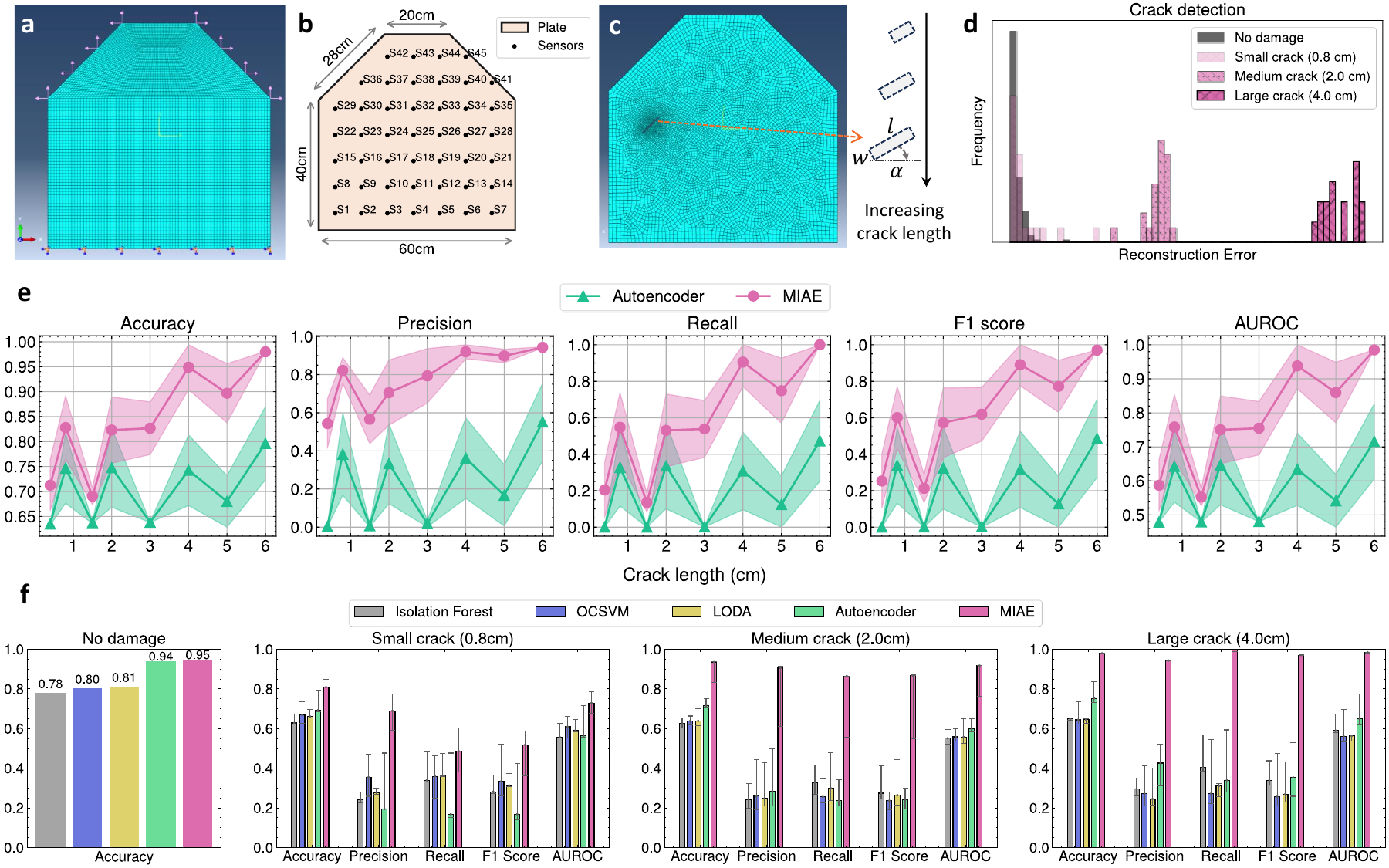}
\caption{\textbf{Damage detection for a cracked gusset plate.} \textbf{a}. Finite element mesh of an intact plate, boundary conditions, and loading. \textbf{b}. Sensor arrangement with labels. \textbf{c}. A typical cracked plate and its meshing. Different crack lengths represent damage progression. \textbf{d}. Distributions of reconstruction errors of the structure from its undamaged reference and damaged states. As the crack progresses (three different crack lengths), the error distribution shifts to the right and becomes more distinct from the undamaged reference. \textbf{e}. Damage detection performance as the crack length increases. \networkName\ outperforms the baseline autoencoder in all five metrics, especially in the early stages of damage emergence. \textbf{f}. Compared to baseline anomaly detection methods, \networkName\ exhibits the best detection accuracy in the undamaged scenario and consistently achieves higher damage detection rates across all the evaluated metrics and crack lengths.\label{fig:2} }
\end{figure*}

\begin{figure*}[!ht]
\centering
\includegraphics[width=0.9\textwidth]{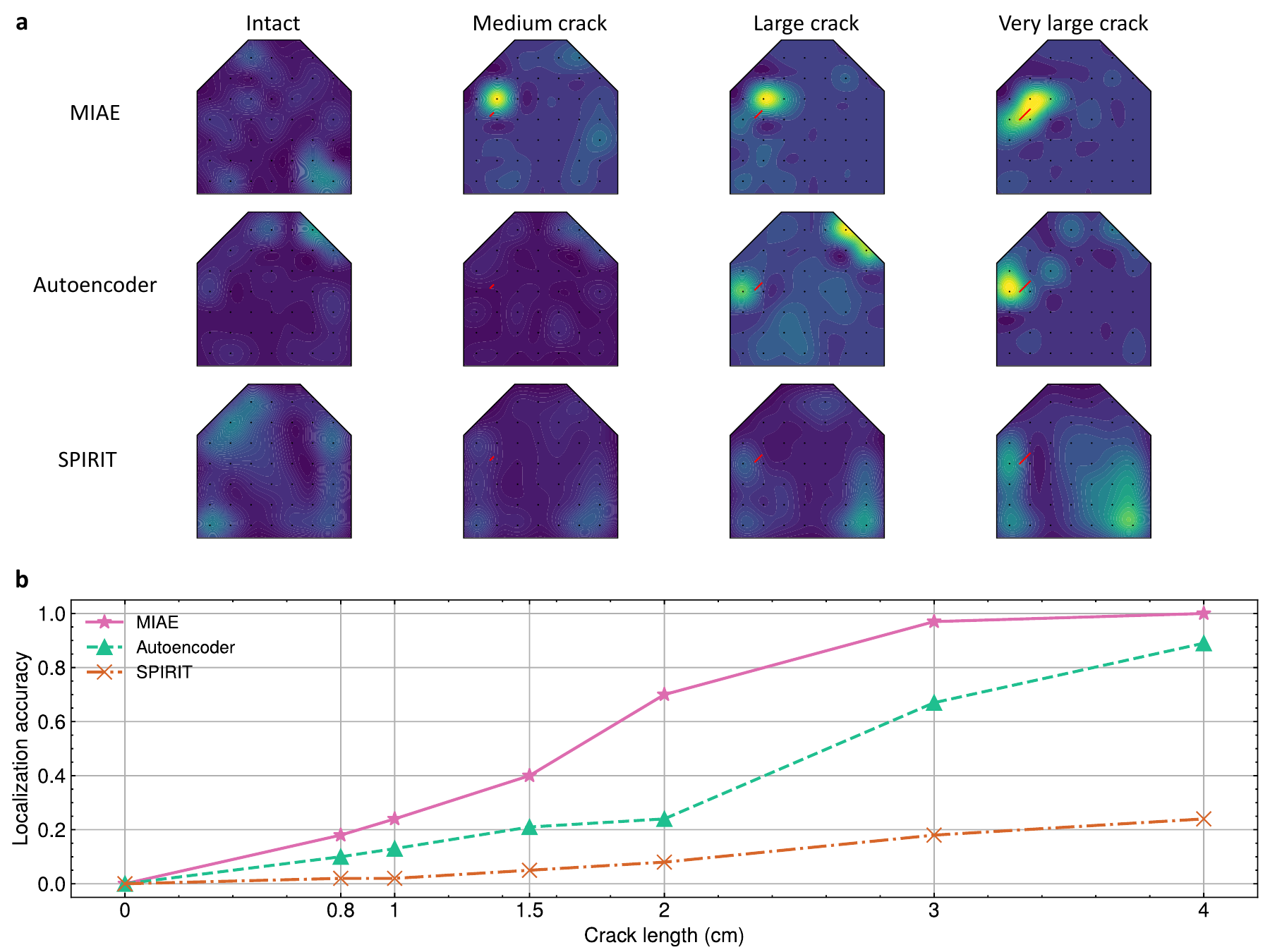}
\caption{\textbf{Damage localization for a cracked gusset plate.} We consider different crack lengths: intact (0$cm$), medium (2$cm$), large (4$cm$), and very-large (6$cm$). \networkName\ localizes cracks at an earlier damage stage than prior unsupervised methods. \textbf{a}. Damage score maps for different damage scenarios. A high damage score (peak values in yellow) at one or more sensors near the crack indicates successful localization. \networkName\ can localize the crack earlier (at a small crack length) than SPIRIT and autoencoder. \textbf{b}. Damage localization accuracy from an extensive analysis of 37 different crack scenarios. The $y$-axis refers to the percentage of cases where damage was successfully localized. Compared to autoencoder and SPIRIT, \networkName\ has a higher localization accuracy across all crack lengths (e.g., 35\% better localization for 2$cm$ long cracks), demonstrating its ability to localize cracks earlier than the baseline approaches. \label{fig:3}
}
\end{figure*}

\vspace{2pt}
\textbf{Numerical simulation--a gusset plate.} An intact (undamaged) polygon-shaped steel plate is analyzed using finite element simulations. The mesh details are shown in Fig. \ref{fig:2}a. This undamaged plate is subjected to random traffic loads to simulate the normal operations of a structural component. The detailed dimensions of the plate are shown in Fig. \ref{fig:2}b (thickness is $1.2cm$). Strain responses are measured at 45 points within the structure as marked in Fig. \ref{fig:2}b.

\vspace{2pt}
\emph{Establishing reference baseline of structural behavior:} Time-series data from the sensors are measured, segmented, and then compressed (more details are provided in the Methods section). The compressed data consists of the running mean $\mu$ and standard deviation $\sigma$ for each sensor. Subsequently, the \networkName\ utilizes this data for training by seeking to reconstruct the input compressed sensor data. The trained network computes a reference for reconstruction errors, which involves the mean squared error between the input and output for each sensor. This reference is the intact structure's baseline, representing the undamaged structural condition.

\vspace{2pt}
\emph{Damage detection evaluation:} We randomly introduced cracks at various locations in the finite element model. To simulate different cracks in each damage scenario, we increase the crack length $l$ from $0.4cm$ to $6cm$ while keeping the crack width $w$ and orientation (angle) $\alpha$ fixed. An example of the damaged plate and the corresponding mesh is illustrated in Fig. \ref{fig:2}c (\textcolor{black}{$w=0.4cm$ and $\alpha=30$\textdegree}). For evaluation, compressed sensor data is now obtained from the damaged plate over small time windows and processed by the trained model to obtain their reconstruction errors. These errors are compared to the reference reconstruction errors from baseline behavior to identify damage in the structure. The distribution of the reconstruction errors can reveal how closely the response behavior of the damaged structure resembles the original undamaged system. Additionally, we limit the number of anomaly data samples to 20 or fewer to demonstrate \methodName{}'s rapid damage detection capabilities. Such a capability allows \methodName\ to be efficient, effective, and deployable in real-world applications for near-real-time damage detection.

Figure~\ref{fig:2}d presents reconstruction error histograms comparing the undamaged baseline to the damaged structure with varying crack lengths. For small damage ($0.8cm$), the reconstruction errors overlap with the reference undamaged reconstruction errors with only minor separation in the distributions,  suggesting a similarity in structural response behavior. As the damage grows to a crack of length $2cm$, noticeable differences emerge between the two reconstruction error distributions. These disparities indicate that the model cannot accurately reconstruct the sensor data due to the distribution shift and can thus detect the damage. Furthermore, as the crack length increases to $4cm$, the distribution of reconstruction errors for data from the damage shifts towards higher magnitudes. Therefore, damage to the structure can be easily detected in this case.

We also evaluated the proposed \networkName\ against a standard autoencoder w.r.t. a range of metrics, including accuracy, precision, F1-score, and area under the receiver operating characteristic (AUROC)~\cite{bewick2004statistics} (detailed information on computing these metrics is provided in the Methods section). Figure \ref{fig:2}e reveals that \networkName\ outperforms the autoencoders in all five metrics across a wide range of crack lengths. Crucially, \networkName\ exhibits significant improvement in detection performance when the crack is minor (before $2cm$), which is highly desirable for early detection in real applications, especially on fracture-critical structural components that typically lack a baseline model and exhibit large behavioral differences even among similarly designed components.

\vspace{2pt}
\emph{Detection performance comparison against other ML methods:} We compare \networkName\ with four baseline methods: Isolation Forest \cite{liu2008isolation}, One-Class support vector machines (OCSVM), LODA \cite{pevny2016loda}, and autoencoders using sensor data from small ($0.8cm$), medium ($2cm$), and large ($4cm$) crack lengths, across 37 cases with cracks at different locations and widths. The results are shown in Fig.~\ref{fig:2}f. \networkName\ consistently surpasses all other methods in accuracy, recall, F1-score, and AUROC. Compared to standard autoencoders, the incorporated mechanical knowledge in \networkName\ significantly improves damage detection performance, particularly for small cracks.

\emph{Damage localization evaluation:} Apart from detecting damage, another critical desideratum of SHM is localizing the damage on the structure. \networkName\ demonstrates robust localization ability, even when the damage is relatively small. Unlike the detection process, which involves comparing reconstruction errors from all the sensors, localization is performed by computing the norms of reconstruction errors at each sensor to obtain a damage score (see Method section for details). A high score indicates the presence of damage adjacent to that sensor. To localize the damage more precisely, we interpolate the scores between the sensors and identify the peak score location.

Figure \ref{fig:3}a shows the damage localization heatmaps for different crack lengths and the exact damage location (red line). The intact structure exhibits a uniform damage score in the first column, indicating the absence of detected damage. As a crack emerges, \networkName\ accurately localizes a medium-size crack ($2cm$) and a large-size crack ($4cm$), as indicated by a high damage score (yellow region). The high damage score precisely overlays the cracked region as it grows to a very large crack ($6cm$).

Here, we compare against two baseline dimensionality reduction methods, (i) SPIRIT~\cite{papadimitriou2005streaming, shafer2012rainmon}, which performs linear dimensionality reduction through online PCA, and (ii) a standard autoencoder that performs non-linear dimensionality reduction through a deep neural network. Compared to the baselines (second and third row of Fig.~\ref{fig:3}a for SPIRIT and autoencoder, respectively), \networkName\ (non-linear dimensionality reduction with mechanical consistency) is capable of localizing damage at an earlier stage  ($2cm$, second column) of crack propagation. Autoencoder can only localize the very-large crack (6$cm$, fourth column), while SPIRIT completely fails to localize the crack. These results highlight the benefit of non-linear (autoencoder) over linear (SPIRIT) dimensionality reduction and the additional benefit afforded by incorporating mechanical constraints (\networkName{}).

% It is also worth noting that when the crack is small ($0.8cm$, second column), a higher damage score near the structural boundaries may not conclusively identify the correct crack location since structural response variations in those areas are typically larger. But, the damage scores are still higher than the undamaged reference, and the score map still indicates the presence of damage within the structure.

We also evaluated damage localization accuracy for the same damage detection cases we considered earlier. Figure \ref{fig:3}b reports the fraction of cases, out of 37, where the damage was successfully localized at different crack lengths. Compared to autoencoder and SPIRIT, \networkName\ has an overall higher success rate and around 35\% better localization for medium-sized cracks ranging from 1.5 to 3$cm$. Furthermore, \networkName\ can localize most of the cracks at a size of 3$cm$ while the autoencoder still fails in many cases. Damage localization results for other cases are shown in Fig. 1 of the Supplementary information, and Movie 1 of the Supplementary shows a visualization of the damage detection and localization process.

\begin{figure*}[!ht]
\centering
\includegraphics[width=0.98\textwidth]{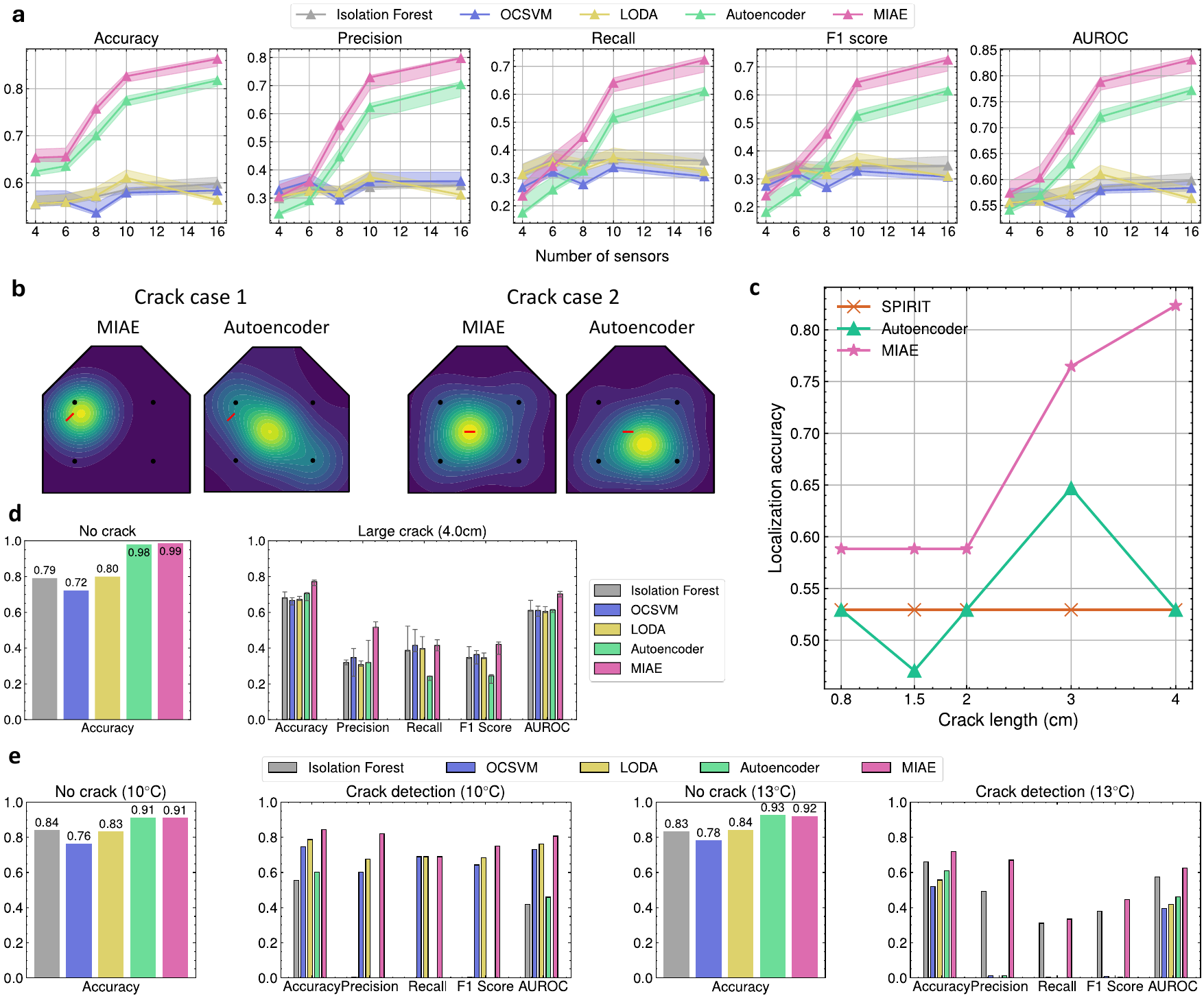}
\caption{ \textcolor{black}{\textbf{Damage detection and localization under sensor and temperature variations.} \textbf{a}. Damage detection performance as the number of sensors varies. \textbf{b}. Comparison of localization accuracy between \networkName\ and autoencoder with four sensors for two different crack scenarios. \networkName’s peak damage score is closer to the true crack location in both cases. \textbf{c}. Comparison of damage localization accuracy with four sensors as crack length increases. \networkName\ outperforms the baseline approaches. \textbf{d}. Damage detection performance with noisy (0.5\% additive Gaussian noise) sensor data. \textbf{e}. Damage detection performance was evaluated at two different temperatures.}\label{fig:4}}
\end{figure*}

\vspace{2pt}
\textcolor{black}{\emph{Damage detection and localization with reduced number of sensors:} So far, we evaluated the damage detection and localization performance of \networkName\ using all available sensors (45 in number). However, real-world applications seek to minimize the number of sensors and instead place a few sensors strategically. Therefore, we evaluate the damage detection and localization performance by varying the number of sensors. When the number of sensors is fewer than 10, they are strategically selected to ensure coverage over the plate (see Supplementary note 6 for details). Otherwise, the sensors are placed randomly on the structure. To ensure reliability, we repeated the evaluation multiple times for a given sensor budget, each time with a different configuration. Figure \ref{fig:4}a shows the damage detection performance for a crack size of 0.8$cm$ as we vary the number of sensors. The performance of methods such as Isolation Forest, OCSVM, and LODA shows no appreciable improvement as we increase the number of sensors since they are designed to operate separately on data from each sensor. In contrast, autoencoder and \networkName\ are learned on data available from all sensors. They can better leverage the additional information available as we increase the number of sensors and thus gain performance. Importantly, \networkName\ leverages sensor correlations based on mechanics knowledge, achieving the best performance among all evaluated methods with only four sensors while getting more accurate as more sensors are available.} 

\textcolor{black}{Figure \ref{fig:4}b shows a configuration of four sensors (S9, S13, S30, and S34, marked as black dots within the localization map) utilized to localize damage from different scenarios. Compared to the standard autoencoder, \networkName\ achieves better localization accuracy (notice that the peak damage scores are closer to the crack). SPIRIT failed to localize damage with only four sensors, so we do not report these results. Next, we extensively analyze the localization performance as the fraction of cases correctly localized as the crack size increases. Specifically, in the 4-sensor setup, we estimate the peak damage score location as the centroid of the four sensors, which is weighted by their damage scores. In this case, we define localization as successful if the true damage is within a radius of 13$cm$ (half of the sensor-to-sensor gap) around the peak location in the damage score map. As shown in Fig. \ref{fig:4}c, \networkName\ outperforms both the autoencoder and SPIRIT, achieving around 10\% to 35\% better localization performance across different crack lengths. In summary, even with a limited number of sensors, \networkName\ exhibits excellent damage detection and localization performance.}

\vspace{2pt}
\textcolor{black}{\emph{Environmental effects consideration:} Here, we explore the impact of environmental factors, such as noisy data sources and temperature variations, on structural damage assessment. Since strain sensors typically provide highly accurate measurements, a Gaussian noise level of 0.5\% is introduced to the raw strain data from four sensors (S9, S13, S30, and S34). This data undergoes preprocessing (compression), and \networkName\ is trained on such data from the structures's undamaged state. The trained model is then evaluated using noisy sensor data under various crack scenarios. Figure \ref{fig:4}d shows the testing accuracy for undamaged data and damage detection performance. Even with only four sensors, \networkName\ outperforms the other models when evaluated on undamaged scenarios and excels at damage detection for large cracks of length 4$cm$ (can detect even smaller cracks if more sensors are used). These results underscore \networkName's robustness against noisy sensor data for detecting minor damage, i.e., at an early stage.}

\begin{figure*}[!ht]
\centering
\includegraphics[width=0.9\textwidth]{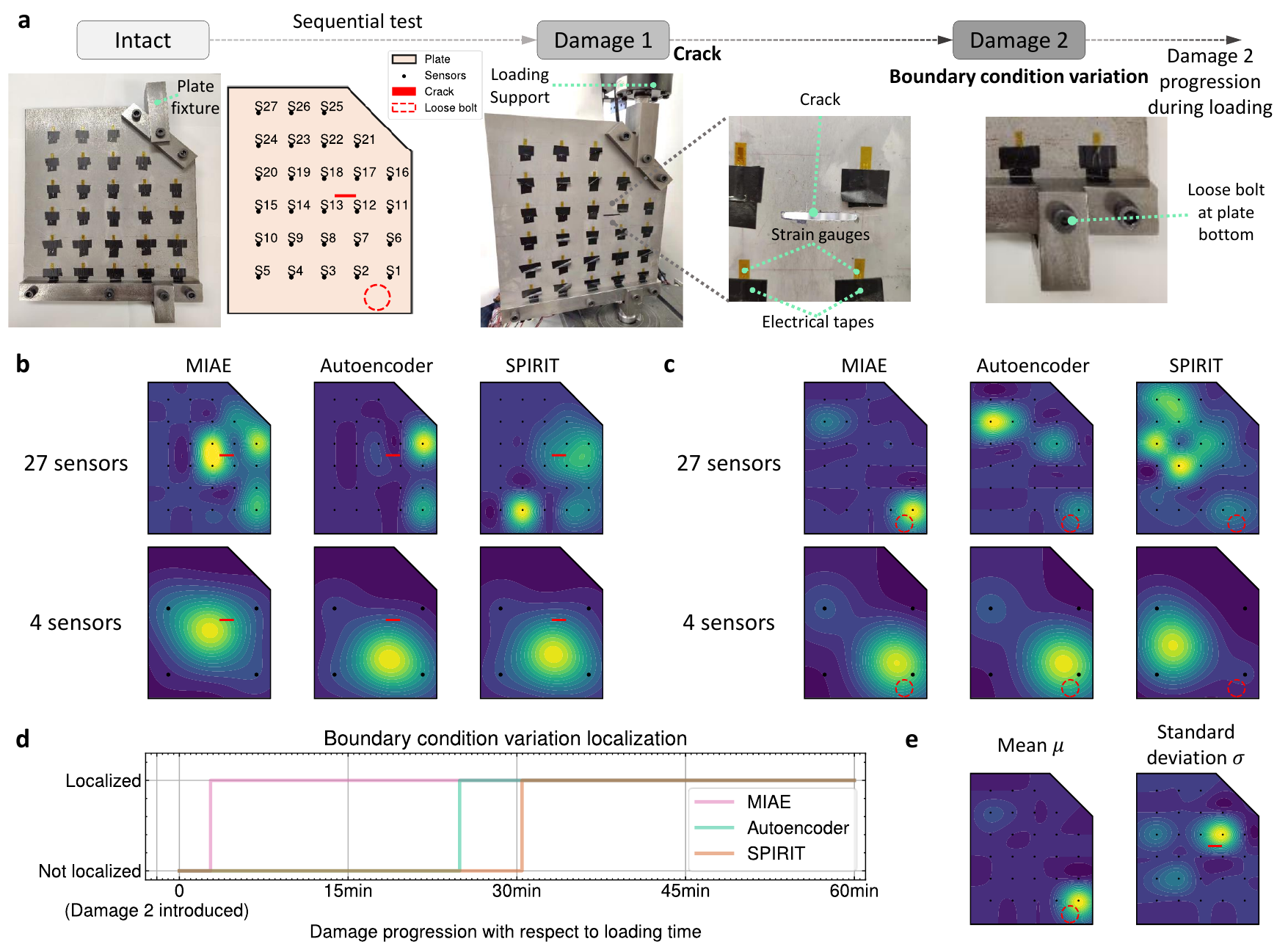}
\caption{\textbf{Laboratory experiment on a steel plate structure.} \textbf{a}. Two types of damage are introduced sequentially (a crack and boundary condition variation). The crack is located in the middle of the plate, and the second damage was introduced by loosening the bolt connections. Under loading, the connection of the plate loosens, thus mimicking damage progression. \textbf{b}. Crack localization results with 27 and 4 sensors, respectively. When using all 27 sensors, \networkName\ accurately delineates the crack region with a high damage score (yellow region) around the crack tips, outperforming the autoencoder. SPIRIT fails to localize the damage in both setups. \textbf{c}. Localization for bolt loosening damage under loading. \networkName\ correctly localizes the damage at the bottom plate connection in the early loading stage (damage progression). \textbf{d}. Localization performance for boundary condition variation. Only \networkName\ can localize the damage early. As the crack size increases, both the autoencoder and SPIRIT gradually succeed in localizing it. \textbf{e}. Damage differentiation through compressed sensor data $\mu$ and $\sigma$. While $\mu$ is more sensitive to boundary condition changes, $\sigma$ responds more to cracks in the structure. \label{fig:5} }
\end{figure*}

\textcolor{black}{We analyze the temperature effect by applying different temperature environments to the structure under loading. The same sensor configuration is utilized as in the noisy data scenario. For training, data from the undamaged structure is measured at temperatures between 5\textdegree$C$ and 30\textdegree$C$ with intervals of 5\textdegree$C$. After training, the model was evaluated at 10\textdegree$C$ and 13\textdegree$C$ for undamaged and damaged cases. Figure \ref{fig:4}e shows damage detection performance for these configurations. Specifically, the autoencoder achieves similar performance as \networkName\ for undamaged cases but fails to detect damages at 10 and 13\textdegree$C$. This demonstrates that incorporating mechanical strain relations between the sensors into the autoencoder increases its robustness to temperature variations.}

\textcolor{black}{LODA and Isolation Forest can obtain comparable recall scores during damage detection evaluation. However, their accuracy is low when evaluating undamaged samples, making these damage detection results less reliable. Overall, \networkName\ outperforms the other baseline methods. These numerical results provide comprehensive coverage across various scenarios, enabling the model to distinguish actual structural damage from effects caused by unknown temperature variations, even if they are not included during training. At last, damage localization is also performed for noisy data scenarios and temperature variations, with results similar to those shown in Fig.~\ref{fig:3}a. We omit these results for brevity.}

\vspace{2pt}
\textbf{Experimental validation--a gusset plate.} We evaluate \methodName\ on a plate structure (Fig. \ref{fig:5}a) to demonstrate its feasibility. The experimented steel plate measures approximately $45cm$ $\times$ $36cm$. Twenty-seven (27) strain sensors were attached to the plate surface with a center-to-center gap of $6.5cm$. Random traffic-like loading is applied to the intact plate structure for 3 hours to generate enough data to train \networkName{}. Then, we introduced damage to the plate. To demonstrate the ability to differentiate between damage types, we sequentially evaluated two typical types of damage---cracks followed by boundary condition variations---applied on the plates during the experiment. This approach allows us to illustrate the progression of damage. Figure~\ref{fig:5}a shows the first damage, a crack of size $4cm$ $\times$ $0.5cm$, introduced in the middle right side of the plate. The second type of damage (boundary condition variations) was subsequently introduced at the lower boundary connection of the plate. The damage was introduced by manually loosening the bolt connecting the plate to the loading frame. The bolt was loosened continuously throughout the experimental loading to mimic the progression of the boundary condition damage. In both damage states, random traffic loading was applied to the plate before and after introducing damage, and corresponding strain response data were recorded from all sensors. Data from damaged structures was evaluated similarly to the finite element simulation. Details of the sensor placement and two damage locations are shown in Fig.~\ref{fig:5}a.

% Figure \ref{fig:5}c shows damage detection results for both scenarios. We compare Isolation Forest, One-class Support Vector Machines (OCSVM), LODA, autoencoders, and \methodName. Results show that, for the large damage we introduced, \methodName{}'s performance is on par with the autoencoder but still outperforms other baseline methods. \methodName{}'s primary advantage over the baselines is detecting minor damages. {\color{blue} VB: which results support this claim?} {\color{blue}VB: ok, remove this figure and paragraph.}

\vspace{2pt}
\emph{Damage Detection and Localization:} When considering the significant damage we introduced, the performance of \networkName\ is comparable to that of the autoencoder. However, \networkName\ can localize the damage more accurately than the autoencoder. Figure \ref{fig:5}b presents the crack localization score maps as we vary the number of sensors. Compared to the standard autoencoder, the integrated mechanical knowledge significantly improves the damage localization accuracy. When using all available sensors, the score map exhibits a much larger peak region on both sides of the crack. This occurs because stress concentration primarily occurs at the crack's tips, and the sensors on both sides of the tip sense the structural response variations. The autoencoder score map exhibits a similar pattern, but the peak scores are much lower (faint yellow) near the right side of the crack tip, resulting in very weak localization. SPIRIT completely failed to localize the crack.

\textcolor{black}{When using only four, instead of twenty-seven, sensors, \networkName\ had the best localization performance, with a smaller distance between the peak in the score map and the crack location than autoencoder and SPIRIT.} These results suggest that our proposed method is more sensitive to minor damage, amplifying such discrepancies and improving localization over a standard autoencoder.

Figure \ref{fig:5}c shows the damage score map for the boundary condition variation in the first $2min$ of loading after manually loosening the plate connections (i.e., introducing second damage). Only \networkName\ successfully localized the damage at the bottom right corner of the plate. This region corresponds to the actual location of the boundary variations we introduced. Autoencoder and SPIRIT exhibit worse localization performance with late localization during damage progression. Meanwhile, in the second row of Figure \ref{fig:5}c, we demonstrate that accurate localization can also be achieved with fewer sensors. Figure \ref{fig:5}d shows the progression of damage localization across different methods. Autoencoder and SPIRIT can only localize the damage after $20min$ of loading. This again demonstrates that \networkName\ exhibits higher sensitivity, enabling early damage localization after 2.5$min$ of loading. More details of this evaluation can be found in Fig. 2 of the Supplementary. Overall, our results indicate that \networkName\ achieves early detection and localization for different types of damage.

\begin{figure*}[!ht]
\centering
\includegraphics[width=\textwidth]{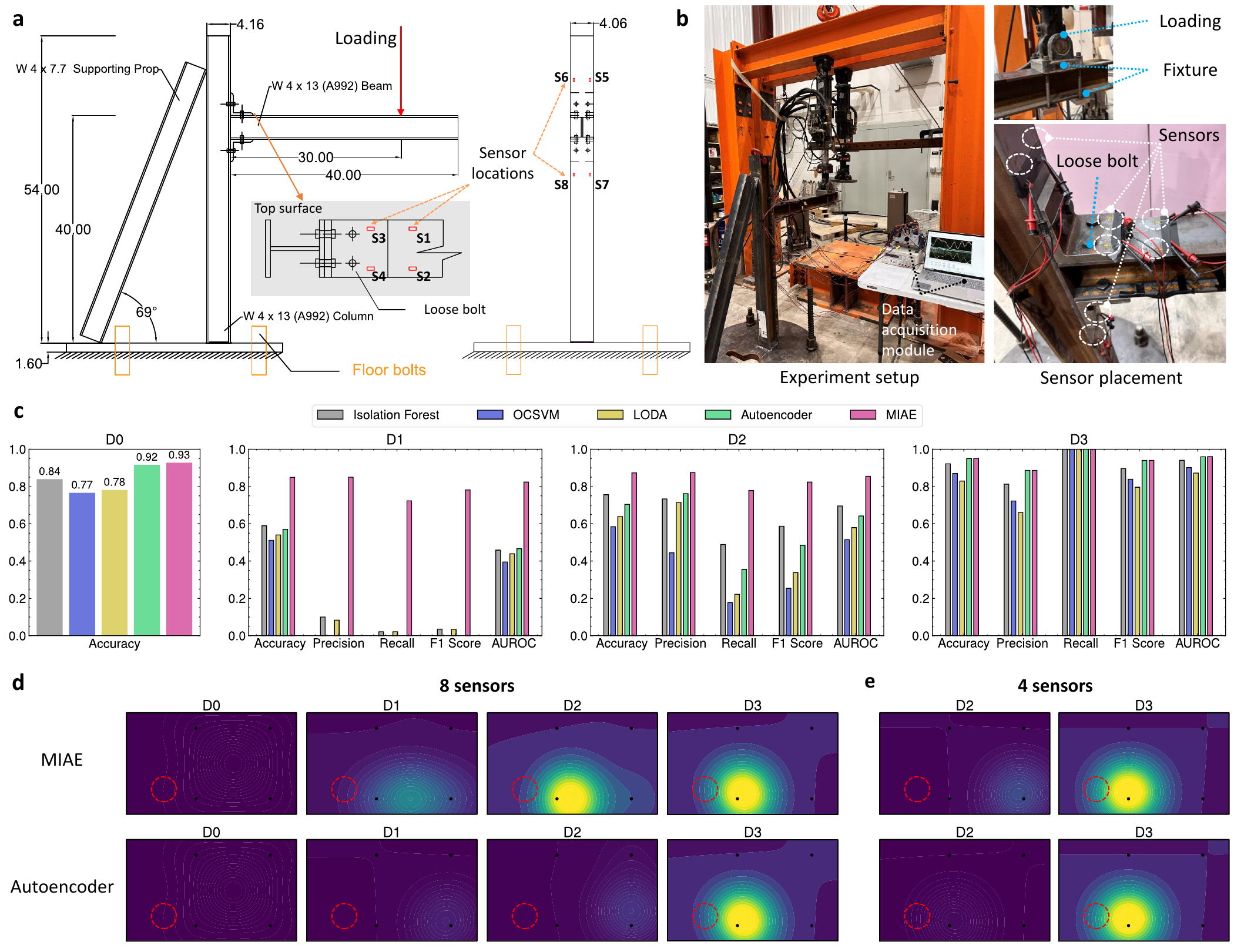}
\caption{\textcolor{black}{\textbf{Laboratory experiment on beam-column structure.} \textbf{a}. The design specifications. \textbf{b}. The experimental setup, sensor placement, and loading details. \textbf{c}. The damage detection performance at different levels of damage. \textbf{d}. Damage localization using all eight sensors' data for training. Only four sensors on the beam and support are used to generate the localization map. \textbf{e}. Damage localization with only four sensors on the support and beam for training and evaluation. Overall, \networkName\ can detect and localize small damage, achieving the best performance among all comparisons to the baselines.}}
\label{fig:6}
\end{figure*}

\vspace{2pt}
\emph{Damage identification:} In addition to damage detection and localization, \networkName\ can also distinguish unseen types of damage based on the compressed sensor data features $\mu$ and $\sigma$. We independently compute the difference of norms for $\mu$ and $\sigma$ for each sensor without combining them in the damage score (see Method section and Equation \ref{eq2:p} for details). Figure \ref{fig:5}e presents two distinct damage score maps derived from $\mu$ and $\sigma$. We observe that $\mu$, which represents the temporal average of the strain responses, is more sensitive to boundary condition variations. This behavior is attributed to the fact that the loosening of structural connections reduces the structure's stiffness, resulting in an overall attenuation in strain magnitudes. On the other hand, the standard deviation $\sigma$ is more sensitive to the cracks induced by stress concentrations. Sensors positioned near the crack tips experience elevated strain during loading, leading to a larger deviation from the baseline strain responses, i.e., increased standard deviation. We show additional results on distinguishing different types of damage in Fig. 4 of the Supplementary.

\vspace{2pt}
\textcolor{black}{
\textbf{Experimental validation--a beam-column structure.} Beam columns are structural elements frequently encountered in various engineering applications, such as building frames, bridges, and industrial structures. The widespread use of beam columns in structures highlights their importance in structural engineering and the need for engineers to understand their behavior and design principles thoroughly.}

\textcolor{black}{We consider a structure with multiple connected components. Figure \ref{fig:6}a illustrates our setup consisting of a column, beam, and other components, with units in inches. The load is applied at 3/4 of the span of the beam, 76.2$cm$ (30 inches) from the column face. Strain sensors are strategically placed at the support, beam, and column flange. Figure \ref{fig:6}b shows a picture of our experimental setup in the lab and the loading details.}

\textcolor{black}{After loading the undamaged structure (state D0), we introduced different levels of damage in the form of variations in boundary conditions (bolt loosening) during loading. The bolt near sensor four (S4) is progressively loosened from an intact state of 80 $lb\cdot ft$ to around 60 $lb\cdot ft$ for three levels of damage (D1, D2, and D3).}

\vspace{2pt}
\textcolor{black}{\emph{Damage Detection:} Time-series strain data was recorded for the entire experiment and compressed for model training and evaluation. Figure~\ref{fig:6}c shows the detection accuracy for the intact structure. In the undamaged state D0, \networkName\ achieves the highest accuracy compared to other baseline methods, indicating excellent learning of the undamaged reference state. When detecting damage, \networkName\ demonstrates superior performance at the early stages of levels 1 and 2. At damage level 3, almost all methods can detect damage. However, methods like OCSVM and LODA achieve low accuracy when evaluating the undamaged case, making their damage detection results less reliable. Overall, the results demonstrate \networkName's ability to detect damage earlier than existing methods.}

\vspace{2pt}
\textcolor{black}{\emph{Damage Localization:} We compute the damage scores at all eight sensors for different damage levels. Sensors on the beam and support (S1-S4) exhibit relatively higher damage scores compared to the other four sensors in the column, indicating damage nearby. We use only these sensors to compute damage score maps and localize the damage as they are in-plane with the beam, while the others are not. In Fig. \ref{fig:6}d, when the first level of damage (D1) occurs, \networkName\ has its peak damage score between sensors 2 and 4, indicating potential damage. Still, it does not accurately localize the damage near sensor 4. But at damage states D2 and D3, \networkName\ accurately localizes the damage near sensor 4. The baseline autoencoder only localizes the damage at D3, while SPIRIT fails to localize any damage (results omitted for brevity). These results demonstrate that \networkName\ enhances localization for low-severity damage.}

\textcolor{black}{It is noteworthy that although sensors at the column are not directly used in the localization, they greatly contribute to the model training and establish the reference baseline behavior of the structure. To illustrate this, we consider a configuration using only four sensors located on the beam and the support (excluding the sensors on the column). In Fig.~\ref{fig:6}e, \networkName\ can hardly localize damage at the second damage level D2. And both \networkName\ and autoencoder can localize the damage at D3. Compared to successful localization at D2 when using eight sensors with \networkName, this delayed localization using fewer sensors demonstrates that additional sensors on the other structural component greatly enhance \networkName's performance.}

\section{Discussion}
This paper presented \methodName\ for automated detection and localization of unforeseen damage, as well as the differentiation between different types of damage. \methodName\ leveraged sensors positioned at various locations to gather time-series data from an intact structure, which were compressed into features at each sensor and employed for training a mechanics-informed autoencoder. The overall idea of \methodName\ was to learn a reference model of strain responses from an intact structure, which aids in capturing anomalies indicative of structural damage. We demonstrated the efficacy of \methodName\ through both numerical and laboratory experiments on two structures, namely, a gusset plate and a beam column structure. 

A key component of \methodName{} is the Mechanics-informed autoencoder (\networkName{}). It leveraged the relationships between sensors based on their mechanical strain responses to enhance detection during early damage progression and enable earlier damage localization than prior methods. \networkName\ is sample efficient, requiring only a minimal amount of data samples for damage detection and localization. \networkName\ outperformed standard machine learning techniques like One-Class SVM, Isolation Forest, and LODA in detecting damage across different damage scenarios, achieving better accuracy, precision, recall, F-score, and AUROC. Notably, the novel loss function incorporating pairwise mechanical relations between the sensors improved the localization rate of minor damages by up to 35\% over a standard autoencoder. In our laboratory experiment on a steel plate, \methodName\ could also distinguish between different types of damage (boundary condition variations and cracks). \textcolor{black}{Finally, the experiment on a beam and column structure demonstrated the generalization ability of \methodName\ to complex structures with multiple components and different geometries.}

\textcolor{black}{The data compression technique used in this work has been previously developed by our research group to achieve low-cost field deployable edge computing on ultra-low-powered wireless sensors. The method has been validated in laboratory and field tests \cite{alavi2016intelligent, alavi2016damage, hasni2017new, hasni2017continuous}. This work's application enhances and distinctly sets our method apart from existing autoencoder-based techniques that typically process raw time-series signals directly. Incorporating data compression affords robustness to sensor noise and enables more efficient data processing, network training, and prediction, facilitating near-real-time damage detection and localization. This is extremely important for advanced wireless sensors that require efficient data storage and transmission. Overall, this technology underpins the practicality of our approach in real-world applications, contributing to an efficient and automated SHM solution.}

We demonstrated the utility of \methodName\ as a SHM framework for near-real-time detection and localization of structural damage. We evaluated it across various numerical and laboratory experiments, including gusset plate structures and a large-scale beam-column structure involving multiple connected components. An exciting direction of future work can focus on scaling \methodName\ to even larger-scale structures (e.g., entire bridge or building). This would necessitate optimizing sensor placement, using a heterogeneous suite of sensors, and adapting the mechanics correlations for larger structures and different types of sensors.

\section{Methods}
\textbf{Finite element analysis (FEA).} The gusset plate is simulated using 3D elements (C3D8R) in ABAQUS under clamped-clamped boundary conditions at the bottom edge of the plate. The Poisson ratio and Young’s modulus are 0.32 and 200$GPa$, respectively. To simulate traffic loading, random loading magnitudes are applied to the top left and top right edge in both $-x$ and $-y$ directions. The loading magnitudes are periodic data generated by 100 combinations of Sine and Cosine functions. Each function has a different combination of random phase, frequency, and peak amplitude drawn from normal distributions (see details in Supplementary note 5).

To generate enough training data, the FEA of the undamaged plate structure is repeated with different random loads for multiple iterations. \textcolor{black}{The FEA model uses a fixed timestep of 0.025s.} In the case of damaged structures, random cracks are introduced within the plate geometry, varying in location, length ($l$), width ($w$), and angle ($\alpha$). \textcolor{black}{We varied the crack width from 0.1$cm$ to 0.5$cm$ across different cases, with an interval of 0.1$cm$. And we introduced the crack at an angle of 0, 30, 45, 60, and 90\textdegree}. The mesh size is set to $0.2cm$ for the crack area and $1cm$ for all other regions. The strain responses are obtained by averaging the values across all elements within the specified sensor regions. \textcolor{black}{The strain data in the $y$ direction are recorded at every timestep, with an interval of 0.025s. To analyze the temperature effect, the expansion coefficient of the structure is set to 11$\times10^{-6}$\textdegree$C^{-1}$. The default initial temperature is set to 25\textdegree$C$. The training data is generated at different temperatures varying from 5 to 30\textdegree$C$, with an interval of 5\textdegree$C$}.

\textbf{Laboratory experiment setup.} Young's modulus of the steel plate is unknown for the laboratory experiment of the gusset plate. The strain gauge type is 1-LY11-6/350 and is attached vertically to measure the strain in the vertical direction, aligning with the vertically applied loading. The clamped-clamped boundary conditions of the plate are considered due to their higher controllability in an experimental setup compared to other types of boundary conditions, such as pinned-pinned or pinned-clamped. 

\textcolor{black}{For the beam and column structure, the experimental setup is intended to test the behavior of a beam-column connection with a supporting prop under loading. A moment connection is established between a $W4\times13$ I-beam and a $W4\times13$ column, both made of A992 grade steel. The beam, measuring 40 inches from the column face to its end, is connected to the column using two $L4\times4\times1/2$ web cleats made of A992 grade steel. The cleats are bolted to the beam flange and column flange using two 1/2 inch diameter A325 bolts per cleat, with bolt holes positioned 1 inch from each edge. The spacing between bolts and the edge distance satisfies the minimum edge distance and spacing requirements specified by the AISC manual. The column is connected at the bottom to a circular plate of 36-inch diameter by 3/16-inch fillet welds while it is supported by a $W4\times7.7$ I-section prop at the top by a 1/4-inch fillet weld. The base plate is anchored to the foundation using four anchor bolts. The loading profile is similar to the gusset plate experiment and controlled with a maximum displacement of 0.23 inches, corresponding to a maximum load of approximately 2000 lbs. }

We generated a continuous randomly simulated traffic effect loading profile in both experiments with a time step of $0.1s$. Displacement-control testing was performed using an MTS loading frame model, applying the loading at the top and bottom fixtures. The strain sensors were connected to an NI-9236 strain input module for strain responses monitored during the loading stage, and we collected the raw strain data through LabVIEW. For data compression, we selected seven threshold levels ranging from 30 to 175 $\mu\varepsilon$ with an increment of 24 $\mu\varepsilon$.

\vspace{3pt}
\textbf{Data compression and dataset construction.} In this study, it is assumed that $N$ sensors have been affixed to the structure of interest at $N$ locations. During normal operation, the structure experiences continuous loading forces of unknown magnitude. Each sensor $S_i$ continuously measures a strain signal $\varepsilon_i$ over time (where $i=1, 2, ..., N$). 

\begin{figure*}[t]
\centering
\includegraphics[width=\textwidth]{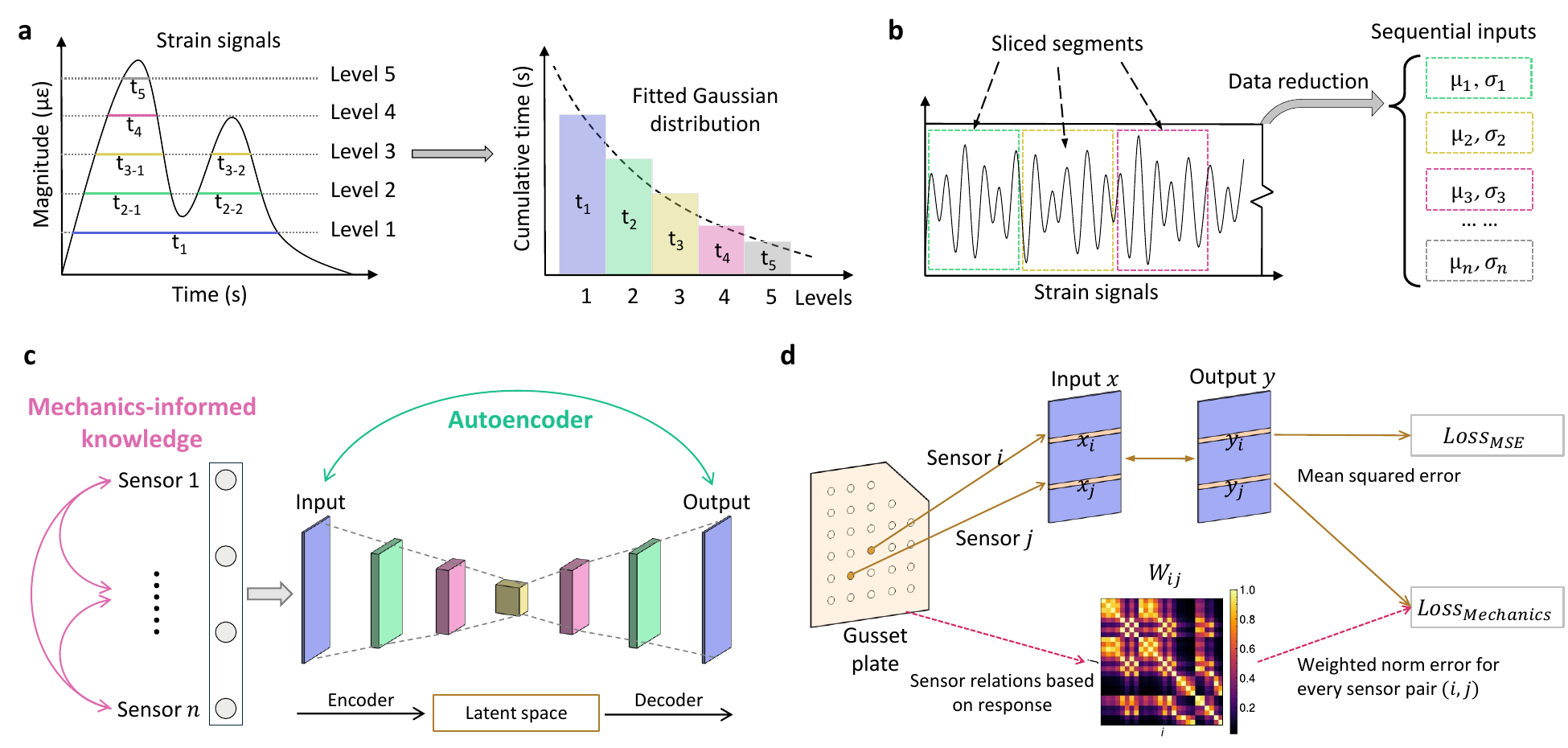}
\caption{\methodName\ methodology. \textbf{a}, Sensor data compression algorithm based on Gaussian distribution. Time-series sensor responses (in micro-strain) from structures are recorded in the left graph, and different threshold levels are defined based on the overall response magnitudes. Next, cumulative events above each threshold level are computed and plotted in the second graph, which is supposed to follow a Gaussian distribution. The best combination of $\mu$ and $\sigma$ is obtained as the compressed sensor data through curve fitting. \textbf{b}, Sensor data processing and dataset construction. \textbf{c}, Our autoencoder architecture. \textbf{d}, The proposed loss function. The weight matrix is computed based on the strain responses from each pair of sensors. Values are shown as contours in the lower part of the graph. \label{fig:7} 
}
\end{figure*}

The data reduction approach is mainly adopted from \cite{bolandi2019novel, hasni2018damage} to solve significant data problems typically generated from structural monitoring sensors (see Fig. \ref{fig:7}a). The approach can be summarized as follows: (i) predefining several strain thresholds based on the overall strain events, (ii) calculating the cumulative times for a selected segment of strain-time responses for all levels of the threshold, (iii) fitting the cumulative time data to the Gaussian equation \ref{eq2:CDF} for data compression, and (iv) obtaining the parameters for Gaussian cumulative density function (CDF) through equation \ref{eq2:CDF}.

\begin{equation}\label{eq2:CDF}
F_{Guassian}(\varepsilon) = \frac{A}{2} \Big[ 1-erf\Big( \frac{\varepsilon-\mu}{\sigma \sqrt{2}} \Big)\Big]
\end{equation}
\noindent where $A$ is the summation of all cumulative time events. $\mu$ and $\sigma$ represent the mean and standard deviation of the cumulative density function, and $erf$ denotes the Gauss error function. To determine the thresholds in Fig. \ref{fig:7}a, the mean strain value $\varepsilon_{mean}$ is computed by averaging the strain responses collected from all sensors in the undamaged structure. Then, the seven threshold values are evenly distributed between $0.5\varepsilon_{mean}$ and $3\varepsilon_{mean}$. For each sensor, every 200 data points are compressed into one set of $\mu$ and $\sigma$ (see Fig. \ref{fig:7}b). 

Next, the compressed sensor data $\mu$ and $\sigma$ (see right side of Fig. \ref{fig:7}b) are utilized to construct the dataset in batches. We use a moving window with length $l=12$ and a stride of 2 to create one batch. For example, the first training sample is taken from the $1^{st}$ to the $12^{th}$ segment, the second training sample is from the $3^{th}$ to the $14^{th}$ segment, then from the $5^{th}$ to the $16^{th}$ segment. The constructed dataset has a size of $B \times l \times 2N$ representing the number of batches, time-series data length, and the number of sensor parameters $\mu$ and $\sigma$, respectively.

\vspace{3pt}
\textbf{Mechanica-Informed Autoencoder network.} Figure \ref{fig:7}c illustrates the proposed mechanics-informed autoencoder architecture with six layers. Specifically, the input and output have the same matrix size, with the output intended to reconstruct the input values. \textcolor{black}{The input layer size is twice the number of sensors employed. For instance, in the case of our numerical simulation, where data from all 45 sensors is used, the input size is 90. In contrast, the size of the middle hidden layers is compacted to 32. In contrast, when using fewer sensors, such as only 4, the size of the middle hidden layer is scaled up by a factor of 8; specifically, each of the first three layers is scaled up by a factor of 2, and the last three layers are scaled down by a factor of 2. This scaling adjustment is necessary because reducing the size of the middle hidden layers beyond this point would not contribute further to model learning. The standard autoencoder has the same architecture as the \networkName\ for all comparisons.} The first part of the loss function is computed as the mean squared error (MSE).

\begin{equation}\label{eq2:loss_MSE}
\mathcal{L}_{MSE} = \frac{1}{n}\sum(\textbf{y}-\textbf{x})^2
\end{equation}
\noindent where $n$, $\textbf{x}$, and $\textbf{y}$ denote the number of samples, the input, and the output from the neural network, respectively.

\textcolor{black}{Most importantly, compared to standard autoencoders, \networkName\ utilizes mechanics-informed knowledge between sensors, leveraging a “mechanics-featured pattern”--inherent in intact structures but absent in damaged ones. This pattern is discerned by analyzing variations in data across different sensors, allowing the model to learn and recognize deviations from the baseline more effectively when damage occurs. Compared to autoencoder, the training reconstruction errors are reduced, while the reconstruction errors on other data for structures usually increased, improving \networkName{}'s sensitivity to subtle damage.} The mechanical characteristics can be incorporated into the neural networks by considering the sensors’ mechanical responses using a weight matrix $W$. \textcolor{black}{Specifically, the matrix has a shape of $N\times N$, and the weight elements are assigned based on corresponding sensor measurements (the largest strain values). This assignment accounts for the correlation of strain changes between two adjacent points in an undamaged structure, effectively reflecting the mechanical features such as the stress concentration effect at boundaries. When accounting for the effect of temperature, the measurements from sensors proximate to the center of the plate are scaled down to one-third before calculating the weight matrix $W$ (as per Equation \ref{eq:w}), and the corresponding $\lambda$ value is reduced by half. These adjustments improve training.}

\textcolor{black}{Furthermore, it is essential to highlight that weights are assigned based on corresponding sensor measurements rather than relying on manual input or predefined assumptions about sensor importance. This approach properly reflects the actual mechanical features of the structure, such as stress concentrations at boundaries, whereas methods that assign weights based on geometry cannot handle them. This capability to utilize raw sensor data to automatically capture and leverage structural mechanics is a crucial aspect of our novel approach.} As shown in Fig. \ref{fig:7}d, the mechanical loss is evaluated for every pair of sensors $i$ and $j$. The mechanics loss term $\mathcal{L}_{Mechanics}$ and the proposed loss function $\mathcal{L}$ can be calculated using the following equations.

\begin{equation}
\mathcal{L}_{Mechanics} = \sum_{i,j}^{N}W_{ij}{(\Delta_i - \Delta_j)}^2
\end{equation}
\begin{equation} \label{eq:norm-delta}
\Delta_i = \parallel\textbf{y}_i\parallel_2^2-\parallel\textbf{x}_i\parallel_2^2
\end{equation}
\begin{equation}\label{eq:w}
W_{ij} = 
\begin{cases}
    max(\varepsilon_i)/max(\varepsilon_j), & \text{if } max(\varepsilon_i) < max(\varepsilon_j) \\
    max(\varepsilon_j)/max(\varepsilon_i), & \text{if } max(\varepsilon_i) \geq max(\varepsilon_j)
\end{cases}
\end{equation}
\begin{equation}\label{eq2:loss_all}
\mathcal{L} = \mathcal{L}_{MSE}+\gamma \mathcal{L}_{Mechanics}
\end{equation}
\noindent where $\Delta_i$ refers to the \textcolor{black}{difference of norms} of the input and output at sensor $i$, and $\Delta$ have shapes of $n \times l \times 2N$. $\mathbf{x}_i$ and $\mathbf{y}_i$ represent the corresponding input and output of the neural network from sensor $i$. The norm operation in equation \ref{eq:norm-delta} is computed along the temporal dimension ($2^{nd}$ dimension). $W$ denotes the weight matrix defined based on each sensor's strain responses, with all element values less than or equal to 1. It is worth noting that $W$ is calculated based on the original strain responses. $\gamma$ is the penalty coefficient for mechanics loss term and is fine-tuned to 0.05 in this study. The proposed model, trained on equation \ref{eq2:loss_all}, enhances the characteristics of structural integrity and sensitivity of model prediction. Data from the damaged structure will not follow the original mechanical features from the intact structure, resulting in poor reconstruction by the neural network and higher reconstruction errors.

\vspace{3pt}
\textbf{Damage detection metric.} After training, the model utilizes the training reconstruction errors $\hat{\Gamma}$ as a reference. It compares them to the reconstruction errors $\Gamma$ at test time to identify any deviations in the samples' distribution. Assuming there are $m$ samples from $N$ sensors, the input, output, and reconstruction error would have $m \times N$ values. The reconstruction error $\Gamma$ for each data point is calculated as:
\begin{equation}\label{eq2:reconstruct}
{\Gamma}_i^j = {(y_i^j - x_i^j)}^2
\end{equation}
\noindent where $j=1,2,\dots,m$ and $i=1,2,\dots,N$. 

To assess the damage detection performance, all samples $\Gamma$ (size of $m \times N)$ are first categorized as either anomaly (positive) or normal data (negative). This classification is accomplished by setting adaptive thresholds based on false positive rates (FPR) derived from training reconstruction errors. Next, we define a ratio $q$ to ascertain whether a testing sample originates from a damaged structure across all $m$ samples. Specifically, if more than $q*N$ of the $N$ sensors were classified as anomalies, the sample is deemed to originate from a damaged structure. As a result, all $m$ samples predict whether the structure is damaged, providing the feasibility of calculating various metrics later on. Due to limited testing data, SMOTEENN~\cite{chawla2002smote, wilson1972asymptotic} was employed to handle class imbalance. Subsequently, sample predictions are compared to the ground truth using binary classification metrics, including accuracy, precision, recall, F1-score, and AUROC.

\vspace{3pt}
\textbf{Damage localization metric.} Damage can be accurately localized by comparing the obtained norm error $\Delta$ from equation \ref{eq:norm-delta} across different sensors. The objective is to summarize the damage condition at each sensor into a single scalar value, and this computation is divided into two steps. First, $\hat{\Delta}$ calculated from the undamaged data and $\Delta$ calculated from damaged data is compared with the reference, considering each sensor parameter $\mu$ or $\sigma$. This intermediate-term is denoted $\mathcal{T}$ as shown in equation \ref{eq2:T}, representing the relative change in reconstruction errors. Second, $\mathcal{T}$ from two types of parameters ($\mu$ and $\sigma$) are integrated as a single metric for conciseness. Therefore, the damage score $p$ is introduced as a damage estimation metric in equation \ref{eq2:p}.

\begin{equation}\label{eq2:T}
\mathcal{T} = \frac
{\Big\vert\frac{1}{m}\sum{\Delta} - \frac{1}{n}\sum{\hat{\Delta}}\Big\vert}
{\frac{1}{n}\sum{\parallel x \parallel}^2_2}
\end{equation}
\begin{equation}\label{eq2:p}
p = \bigg(
\lambda \frac{{\mathcal{T}}^{\mu}}{max({\hat{\mathcal{T}}}^{\mu})} + (1-\lambda) \frac{{\mathcal{T}}^{\sigma}}{max({\hat{\mathcal{T}}}^{\sigma})} 
\bigg)/2
\end{equation}
\noindent where in the first equation, $\frac{1}{m}\sum{\Delta}$ estimates the mean value from all $m$ testing samples, $\frac{1}{n}\sum{\hat{\Delta}}$ represents the mean value from all $n$ training samples, and the denominator $\frac{1}{n}\sum{\parallel x \parallel}$ calculates the average of the corresponding norm. In the second equation, ${\mathcal{T}}^{\mu}$ and ${\mathcal{T}}^{\sigma}$ are vectors of relative parameter errors from parameters $\mu$ and $\sigma$ based on sensors, respectively. $\hat{{\mathcal{T}}}^{\mu}$ and $\hat{{\mathcal{T}}}^{\sigma}$ are the corresponding $\hat{\mathcal{T}}$ calculated from the reference. $\lambda$ is a coefficient that leverages the contribution from parameters $\mu$ and $\sigma$ to damage score $p$. In this work, $\lambda$ is set to 0.5 for \textcolor{black}{all numerical simulations and experimental work}. Furthermore, for damage differentiation, the evaluation for each sensor is performed separately for $\mu$ and $\sigma$, as $\frac{{\mathcal{T}}^{\mu}}{max({\hat{\mathcal{T}}}^{\mu})}$ and $\frac{{\mathcal{T}}^{\sigma}}{max({\hat{\mathcal{T}}}^{\sigma})}$.

The proposed estimation function can effectively differentiate the undamaged and damaged regions based on the damage scores. Specifically, a damage score $p$ less than 1 represents the baseline or the undamaged case, while a higher score demonstrates damage around the sensor region. To some extent, the magnitude of $p$ can indicate the damage severity at the corresponding sensor location. However, such a pattern was not consistently observed when small damage occurred. Furthermore, by incorporating the location information from all sensors and the $p$ scores, the analysis can establish score maps for more precise structural damage localization and estimate the overall structural integrity. \textcolor{black}{When using fewer sensors, the weighted centroid is computed based on the obtained $p$ values and the corresponding sensor's position.}

SPIRIT uses incremental PCA to find correlations and hidden variables that summarize the trend and signify pattern changes. The projection coefficients of the first two hidden variables (i.e., $W_1$ and $W_2$ of the PCA weight matrix $W$) are computed for both the training and testing datasets. The damage scores $p$ are calculated as the element-wise Euclidean distance between the training data point ($W_{i,1}^{train}$, $W_{i,2}^{train}$) and test data point ($W_{i,1}^{test}$, $W_{i,2}^{test}$) where $i=1,2,\ldots,n$. The corresponding norm error $\Delta$ for SPIRIT can be calculated as the element-wise Euclidean distance using equation \ref{eq2:spirit-reconstruct}, while the damage score will be computed as in equation \ref{eq2:p} above.
\begin{equation}\label{eq2:spirit-reconstruct}
\Delta = \sqrt{ {(W_{i,1}^{train}-W_{i,1}^{test})}^2 + {(W_{i,2}^{train}-W_{i,2}^{test})}^2 }
\end{equation}

\textbf{Data availability} The numerical simulation and experimental data supporting the findings of this study are available for reference. The data is available at \url{https://github.com/human-analysis/midas-shm}.

\textbf{Code availability} The codes are provided along with the data at \url{https://github.com/human-analysis/midas-shm}.

\textbf{Acknowledgement} This study was partially supported by the National Science Foundation Award CNS 1645783.

\textbf{Author contributions} N.L. and V.B. supervised the study. X.L., N.L., and V.B. conceived the initial idea and contributed to machine learning. X.L. and H.B. performed finite element analysis and the case studies. X.L. performed the simulation for environmental effects. H.B. contributed to the setup of the laboratory experiment for the gusset plate and prepossessed the data. T.S. and X.L. performed all laboratory experiments and data collection. A. J. designed and performed the beam-column experiment. X.L. and M.M. contributed to Supplementary Information. X.L., M.M., N.L., and V.B. wrote the paper.

\textbf{Competing interests} The authors declare no competing interests.

\textbf{Correspondence and requests for materials} should be addressed to \href{mailto:vishnu@msu.edu}{vishnu@msu.edu} and \href{mailto:lajnefni@msu.edu}{lajnefni@msu.edu}.


% \bibliography{mybib}

\begin{thebibliography}{64}%
\makeatletter
\providecommand \@ifxundefined [1]{%
 \@ifx{#1\undefined}
}%
\providecommand \@ifnum [1]{%
 \ifnum #1\expandafter \@firstoftwo
 \else \expandafter \@secondoftwo
 \fi
}%
\providecommand \@ifx [1]{%
 \ifx #1\expandafter \@firstoftwo
 \else \expandafter \@secondoftwo
 \fi
}%
\providecommand \natexlab [1]{#1}%
\providecommand \enquote  [1]{``#1''}%
\providecommand \bibnamefont  [1]{#1}%
\providecommand \bibfnamefont [1]{#1}%
\providecommand \citenamefont [1]{#1}%
\providecommand \href@noop [0]{\@secondoftwo}%
\providecommand \href [0]{\begingroup \@sanitize@url \@href}%
\providecommand \@href[1]{\@@startlink{#1}\@@href}%
\providecommand \@@href[1]{\endgroup#1\@@endlink}%
\providecommand \@sanitize@url [0]{\catcode `\\12\catcode `\$12\catcode `\&12\catcode `\#12\catcode `\^12\catcode `\_12\catcode `\%12\relax}%
\providecommand \@@startlink[1]{}%
\providecommand \@@endlink[0]{}%
\providecommand \url  [0]{\begingroup\@sanitize@url \@url }%
\providecommand \@url [1]{\endgroup\@href {#1}{\urlprefix }}%
\providecommand \urlprefix  [0]{URL }%
\providecommand \Eprint [0]{\href }%
\providecommand \doibase [0]{https://doi.org/}%
\providecommand \selectlanguage [0]{\@gobble}%
\providecommand \bibinfo  [0]{\@secondoftwo}%
\providecommand \bibfield  [0]{\@secondoftwo}%
\providecommand \translation [1]{[#1]}%
\providecommand \BibitemOpen [0]{}%
\providecommand \bibitemStop [0]{}%
\providecommand \bibitemNoStop [0]{.\EOS\space}%
\providecommand \EOS [0]{\spacefactor3000\relax}%
\providecommand \BibitemShut  [1]{\csname bibitem#1\endcsname}%
\let\auto@bib@innerbib\@empty
%</preamble>
\bibitem [{\citenamefont {Pezeshk}\ \emph {et~al.}(2021)\citenamefont {Pezeshk}, \citenamefont {Camp}, \citenamefont {Kashani}, \citenamefont {Akhani} \emph {et~al.}}]{pezeshk2021data}%
  \BibitemOpen
  \bibfield  {author} {\bibinfo {author} {\bibfnamefont {S.}~\bibnamefont {Pezeshk}}, \bibinfo {author} {\bibfnamefont {C.~V.}\ \bibnamefont {Camp}}, \bibinfo {author} {\bibfnamefont {A.}~\bibnamefont {Kashani}}, \bibinfo {author} {\bibfnamefont {M.}~\bibnamefont {Akhani}}, \emph {et~al.},\ }\bibfield  {title} {\bibinfo {title} {Data analyses from seismic instrumentation installed on the {I}-40 bridge},\ }\href@noop {} {\bibfield  {journal} {\bibinfo  {journal} {Tennessee. Department of Transportation}\ } (\bibinfo {year} {2021})}\BibitemShut {NoStop}%
\bibitem [{\citenamefont {Carnahan}(2022)}]{carnahan2022pittsburgh}%
  \BibitemOpen
  \bibfield  {author} {\bibinfo {author} {\bibfnamefont {H.}~\bibnamefont {Carnahan}},\ }\bibfield  {title} {\bibinfo {title} {Pittsburgh bridge collapse emphasizes need for bridge repairs},\ }\href@noop {} {\bibfield  {journal} {\bibinfo  {journal} {Journal of Protective Coatings \& Linings}\ }\textbf {\bibinfo {volume} {39}},\ \bibinfo {pages} {6} (\bibinfo {year} {2022})}\BibitemShut {NoStop}%
\bibitem [{\citenamefont {Capineri}\ and\ \citenamefont {Bulletti}(2021)}]{capineri2021ultrasonic}%
  \BibitemOpen
  \bibfield  {author} {\bibinfo {author} {\bibfnamefont {L.}~\bibnamefont {Capineri}}\ and\ \bibinfo {author} {\bibfnamefont {A.}~\bibnamefont {Bulletti}},\ }\bibfield  {title} {\bibinfo {title} {Ultrasonic guided-waves sensors and integrated structural health monitoring systems for impact detection and localization: A review},\ }\href@noop {} {\bibfield  {journal} {\bibinfo  {journal} {Sensors}\ }\textbf {\bibinfo {volume} {21}},\ \bibinfo {pages} {2929} (\bibinfo {year} {2021})}\BibitemShut {NoStop}%
\bibitem [{\citenamefont {Tang}\ \emph {et~al.}(2021)\citenamefont {Tang}, \citenamefont {Sui}, \citenamefont {Duan}, \citenamefont {Zhang},\ and\ \citenamefont {Yun}}]{tang2021guided}%
  \BibitemOpen
  \bibfield  {author} {\bibinfo {author} {\bibfnamefont {Z.}~\bibnamefont {Tang}}, \bibinfo {author} {\bibfnamefont {X.}~\bibnamefont {Sui}}, \bibinfo {author} {\bibfnamefont {Y.}~\bibnamefont {Duan}}, \bibinfo {author} {\bibfnamefont {P.}~\bibnamefont {Zhang}},\ and\ \bibinfo {author} {\bibfnamefont {C.~B.}\ \bibnamefont {Yun}},\ }\bibfield  {title} {\bibinfo {title} {Guided wave-based cable damage detection using wave energy transmission and reflection},\ }\href@noop {} {\bibfield  {journal} {\bibinfo  {journal} {Structural Control and Health Monitoring}\ }\textbf {\bibinfo {volume} {28}},\ \bibinfo {pages} {e2688} (\bibinfo {year} {2021})}\BibitemShut {NoStop}%
\bibitem [{\citenamefont {Lissenden}\ \emph {et~al.}(2015)\citenamefont {Lissenden}, \citenamefont {Liu},\ and\ \citenamefont {Rose}}]{lissenden2015use}%
  \BibitemOpen
  \bibfield  {author} {\bibinfo {author} {\bibfnamefont {C.~J.}\ \bibnamefont {Lissenden}}, \bibinfo {author} {\bibfnamefont {Y.}~\bibnamefont {Liu}},\ and\ \bibinfo {author} {\bibfnamefont {J.~L.}\ \bibnamefont {Rose}},\ }\bibfield  {title} {\bibinfo {title} {Use of non-linear ultrasonic guided waves for early damage detection},\ }\href@noop {} {\bibfield  {journal} {\bibinfo  {journal} {Insight-Non-Destructive Testing and Condition Monitoring}\ }\textbf {\bibinfo {volume} {57}},\ \bibinfo {pages} {206} (\bibinfo {year} {2015})}\BibitemShut {NoStop}%
\bibitem [{\citenamefont {Chandarana}\ \emph {et~al.}(2017)\citenamefont {Chandarana}, \citenamefont {Sanchez}, \citenamefont {Soutis},\ and\ \citenamefont {Gresil}}]{chandarana2017early}%
  \BibitemOpen
  \bibfield  {author} {\bibinfo {author} {\bibfnamefont {N.}~\bibnamefont {Chandarana}}, \bibinfo {author} {\bibfnamefont {D.~M.}\ \bibnamefont {Sanchez}}, \bibinfo {author} {\bibfnamefont {C.}~\bibnamefont {Soutis}},\ and\ \bibinfo {author} {\bibfnamefont {M.}~\bibnamefont {Gresil}},\ }\bibfield  {title} {\bibinfo {title} {Early damage detection in composites during fabrication and mechanical testing},\ }\href@noop {} {\bibfield  {journal} {\bibinfo  {journal} {Materials}\ }\textbf {\bibinfo {volume} {10}},\ \bibinfo {pages} {685} (\bibinfo {year} {2017})}\BibitemShut {NoStop}%
\bibitem [{\citenamefont {Song}\ \emph {et~al.}(2022)\citenamefont {Song}, \citenamefont {Zhang}, \citenamefont {Chang},\ and\ \citenamefont {Shen}}]{song2022improved}%
  \BibitemOpen
  \bibfield  {author} {\bibinfo {author} {\bibfnamefont {S.}~\bibnamefont {Song}}, \bibinfo {author} {\bibfnamefont {X.}~\bibnamefont {Zhang}}, \bibinfo {author} {\bibfnamefont {Y.}~\bibnamefont {Chang}},\ and\ \bibinfo {author} {\bibfnamefont {Y.}~\bibnamefont {Shen}},\ }\bibfield  {title} {\bibinfo {title} {An improved structural health monitoring method utilizing sparse representation for acoustic emission signals in rails},\ }\href@noop {} {\bibfield  {journal} {\bibinfo  {journal} {IEEE Transactions on Instrumentation and Measurement}\ }\textbf {\bibinfo {volume} {72}},\ \bibinfo {pages} {1} (\bibinfo {year} {2022})}\BibitemShut {NoStop}%
\bibitem [{\citenamefont {Daneshvar}\ \emph {et~al.}(2021)\citenamefont {Daneshvar}, \citenamefont {Gharighoran}, \citenamefont {Zareei},\ and\ \citenamefont {Karamodin}}]{daneshvar2021early}%
  \BibitemOpen
  \bibfield  {author} {\bibinfo {author} {\bibfnamefont {M.~H.}\ \bibnamefont {Daneshvar}}, \bibinfo {author} {\bibfnamefont {A.}~\bibnamefont {Gharighoran}}, \bibinfo {author} {\bibfnamefont {S.~A.}\ \bibnamefont {Zareei}},\ and\ \bibinfo {author} {\bibfnamefont {A.}~\bibnamefont {Karamodin}},\ }\bibfield  {title} {\bibinfo {title} {Early damage detection under massive data via innovative hybrid methods: application to a large-scale cable-stayed bridge},\ }\href@noop {} {\bibfield  {journal} {\bibinfo  {journal} {Structure and Infrastructure Engineering}\ }\textbf {\bibinfo {volume} {17}},\ \bibinfo {pages} {902} (\bibinfo {year} {2021})}\BibitemShut {NoStop}%
\bibitem [{\citenamefont {Bakhary}\ \emph {et~al.}(2010)\citenamefont {Bakhary}, \citenamefont {Hao},\ and\ \citenamefont {Deeks}}]{bakhary2010substructuring}%
  \BibitemOpen
  \bibfield  {author} {\bibinfo {author} {\bibfnamefont {N.}~\bibnamefont {Bakhary}}, \bibinfo {author} {\bibfnamefont {H.}~\bibnamefont {Hao}},\ and\ \bibinfo {author} {\bibfnamefont {A.~J.}\ \bibnamefont {Deeks}},\ }\bibfield  {title} {\bibinfo {title} {Substructuring technique for damage detection using statistical multi-stage artificial neural network},\ }\href@noop {} {\bibfield  {journal} {\bibinfo  {journal} {Advances in Structural Engineering}\ }\textbf {\bibinfo {volume} {13}},\ \bibinfo {pages} {619} (\bibinfo {year} {2010})}\BibitemShut {NoStop}%
\bibitem [{\citenamefont {Betti}\ \emph {et~al.}(2015)\citenamefont {Betti}, \citenamefont {Facchini},\ and\ \citenamefont {Biagini}}]{betti2015damage}%
  \BibitemOpen
  \bibfield  {author} {\bibinfo {author} {\bibfnamefont {M.}~\bibnamefont {Betti}}, \bibinfo {author} {\bibfnamefont {L.}~\bibnamefont {Facchini}},\ and\ \bibinfo {author} {\bibfnamefont {P.}~\bibnamefont {Biagini}},\ }\bibfield  {title} {\bibinfo {title} {Damage detection on a three-storey steel frame using artificial neural networks and genetic algorithms},\ }\href@noop {} {\bibfield  {journal} {\bibinfo  {journal} {Meccanica}\ }\textbf {\bibinfo {volume} {50}},\ \bibinfo {pages} {875} (\bibinfo {year} {2015})}\BibitemShut {NoStop}%
\bibitem [{\citenamefont {Li}\ \emph {et~al.}(2023)\citenamefont {Li}, \citenamefont {Lin},\ and\ \citenamefont {Zhang}}]{li2023real}%
  \BibitemOpen
  \bibfield  {author} {\bibinfo {author} {\bibfnamefont {Z.}~\bibnamefont {Li}}, \bibinfo {author} {\bibfnamefont {W.}~\bibnamefont {Lin}},\ and\ \bibinfo {author} {\bibfnamefont {Y.}~\bibnamefont {Zhang}},\ }\bibfield  {title} {\bibinfo {title} {Real-time drive-by bridge damage detection using deep auto-encoder},\ }in\ \href@noop {} {\emph {\bibinfo {booktitle} {Structures}}},\ Vol.~\bibinfo {volume} {47}\ (\bibinfo {year} {2023})\ pp.\ \bibinfo {pages} {1167--1181}\BibitemShut {NoStop}%
\bibitem [{\citenamefont {Hasni}\ \emph {et~al.}(2017{\natexlab{a}})\citenamefont {Hasni}, \citenamefont {Alavi}, \citenamefont {Jiao}, \citenamefont {Lajnef}, \citenamefont {Chatti}, \citenamefont {Aono},\ and\ \citenamefont {Chakrabartty}}]{hasni2017new}%
  \BibitemOpen
  \bibfield  {author} {\bibinfo {author} {\bibfnamefont {H.}~\bibnamefont {Hasni}}, \bibinfo {author} {\bibfnamefont {A.~H.}\ \bibnamefont {Alavi}}, \bibinfo {author} {\bibfnamefont {P.}~\bibnamefont {Jiao}}, \bibinfo {author} {\bibfnamefont {N.}~\bibnamefont {Lajnef}}, \bibinfo {author} {\bibfnamefont {K.}~\bibnamefont {Chatti}}, \bibinfo {author} {\bibfnamefont {K.}~\bibnamefont {Aono}},\ and\ \bibinfo {author} {\bibfnamefont {S.}~\bibnamefont {Chakrabartty}},\ }\bibfield  {title} {\bibinfo {title} {A new approach for damage detection in asphalt concrete pavements using battery-free wireless sensors with non-constant injection rates},\ }\href@noop {} {\bibfield  {journal} {\bibinfo  {journal} {Measurement}\ }\textbf {\bibinfo {volume} {110}},\ \bibinfo {pages} {217} (\bibinfo {year} {2017}{\natexlab{a}})}\BibitemShut {NoStop}%
\bibitem [{\citenamefont {Esfandiari}\ \emph {et~al.}(2020)\citenamefont {Esfandiari}, \citenamefont {Nabiyan},\ and\ \citenamefont {Rofooei}}]{esfandiari2020structural}%
  \BibitemOpen
  \bibfield  {author} {\bibinfo {author} {\bibfnamefont {A.}~\bibnamefont {Esfandiari}}, \bibinfo {author} {\bibfnamefont {M.-S.}\ \bibnamefont {Nabiyan}},\ and\ \bibinfo {author} {\bibfnamefont {F.~R.}\ \bibnamefont {Rofooei}},\ }\bibfield  {title} {\bibinfo {title} {Structural damage detection using principal component analysis of frequency response function data},\ }\href@noop {} {\bibfield  {journal} {\bibinfo  {journal} {Structural Control and Health Monitoring}\ }\textbf {\bibinfo {volume} {27}},\ \bibinfo {pages} {e2550} (\bibinfo {year} {2020})}\BibitemShut {NoStop}%
\bibitem [{\citenamefont {Sawant}\ \emph {et~al.}(2023)\citenamefont {Sawant}, \citenamefont {Sethi}, \citenamefont {Banerjee},\ and\ \citenamefont {Tallur}}]{sawant2023unsupervised}%
  \BibitemOpen
  \bibfield  {author} {\bibinfo {author} {\bibfnamefont {S.}~\bibnamefont {Sawant}}, \bibinfo {author} {\bibfnamefont {A.}~\bibnamefont {Sethi}}, \bibinfo {author} {\bibfnamefont {S.}~\bibnamefont {Banerjee}},\ and\ \bibinfo {author} {\bibfnamefont {S.}~\bibnamefont {Tallur}},\ }\bibfield  {title} {\bibinfo {title} {Unsupervised learning framework for temperature compensated damage identification and localization in ultrasonic guided wave {SHM} with transfer learning},\ }\href@noop {} {\bibfield  {journal} {\bibinfo  {journal} {Ultrasonics}\ ,\ \bibinfo {pages} {106931}} (\bibinfo {year} {2023})}\BibitemShut {NoStop}%
\bibitem [{\citenamefont {Sohn}\ \emph {et~al.}(2003)\citenamefont {Sohn}, \citenamefont {Farrar}, \citenamefont {Hemez}, \citenamefont {Shunk}, \citenamefont {Stinemates}, \citenamefont {Nadler},\ and\ \citenamefont {Czarnecki}}]{sohn2003review}%
  \BibitemOpen
  \bibfield  {author} {\bibinfo {author} {\bibfnamefont {H.}~\bibnamefont {Sohn}}, \bibinfo {author} {\bibfnamefont {C.~R.}\ \bibnamefont {Farrar}}, \bibinfo {author} {\bibfnamefont {F.~M.}\ \bibnamefont {Hemez}}, \bibinfo {author} {\bibfnamefont {D.~D.}\ \bibnamefont {Shunk}}, \bibinfo {author} {\bibfnamefont {D.~W.}\ \bibnamefont {Stinemates}}, \bibinfo {author} {\bibfnamefont {B.~R.}\ \bibnamefont {Nadler}},\ and\ \bibinfo {author} {\bibfnamefont {J.~J.}\ \bibnamefont {Czarnecki}},\ }\bibfield  {title} {\bibinfo {title} {A review of structural health monitoring literature: 1996--2001},\ }\href@noop {} {\bibfield  {journal} {\bibinfo  {journal} {Los Alamos National Laboratory, USA}\ }\textbf {\bibinfo {volume} {1}},\ \bibinfo {pages} {16} (\bibinfo {year} {2003})}\BibitemShut {NoStop}%
\bibitem [{\citenamefont {Khan}\ \emph {et~al.}(2016)\citenamefont {Khan}, \citenamefont {Atamturktur}, \citenamefont {Chowdhury},\ and\ \citenamefont {Rahman}}]{khan2016integration}%
  \BibitemOpen
  \bibfield  {author} {\bibinfo {author} {\bibfnamefont {S.~M.}\ \bibnamefont {Khan}}, \bibinfo {author} {\bibfnamefont {S.}~\bibnamefont {Atamturktur}}, \bibinfo {author} {\bibfnamefont {M.}~\bibnamefont {Chowdhury}},\ and\ \bibinfo {author} {\bibfnamefont {M.}~\bibnamefont {Rahman}},\ }\bibfield  {title} {\bibinfo {title} {Integration of structural health monitoring and intelligent transportation systems for bridge condition assessment: Current status and future direction},\ }\href@noop {} {\bibfield  {journal} {\bibinfo  {journal} {IEEE Transactions on Intelligent Transportation Systems}\ }\textbf {\bibinfo {volume} {17}},\ \bibinfo {pages} {2107} (\bibinfo {year} {2016})}\BibitemShut {NoStop}%
\bibitem [{\citenamefont {Wang}\ and\ \citenamefont {Cha}(2021)}]{wang2021unsupervised}%
  \BibitemOpen
  \bibfield  {author} {\bibinfo {author} {\bibfnamefont {Z.}~\bibnamefont {Wang}}\ and\ \bibinfo {author} {\bibfnamefont {Y.-J.}\ \bibnamefont {Cha}},\ }\bibfield  {title} {\bibinfo {title} {Unsupervised deep learning approach using a deep auto-encoder with a one-class support vector machine to detect damage},\ }\href@noop {} {\bibfield  {journal} {\bibinfo  {journal} {Structural Health Monitoring}\ }\textbf {\bibinfo {volume} {20}},\ \bibinfo {pages} {406} (\bibinfo {year} {2021})}\BibitemShut {NoStop}%
\bibitem [{\citenamefont {Jiang}\ \emph {et~al.}(2021)\citenamefont {Jiang}, \citenamefont {Han}, \citenamefont {Du},\ and\ \citenamefont {Ni}}]{jiang2021decentralized}%
  \BibitemOpen
  \bibfield  {author} {\bibinfo {author} {\bibfnamefont {K.}~\bibnamefont {Jiang}}, \bibinfo {author} {\bibfnamefont {Q.}~\bibnamefont {Han}}, \bibinfo {author} {\bibfnamefont {X.}~\bibnamefont {Du}},\ and\ \bibinfo {author} {\bibfnamefont {P.}~\bibnamefont {Ni}},\ }\bibfield  {title} {\bibinfo {title} {A decentralized unsupervised structural condition diagnosis approach using deep auto-encoders},\ }\href@noop {} {\bibfield  {journal} {\bibinfo  {journal} {Computer-Aided Civil and Infrastructure Engineering}\ }\textbf {\bibinfo {volume} {36}},\ \bibinfo {pages} {711} (\bibinfo {year} {2021})}\BibitemShut {NoStop}%
\bibitem [{\citenamefont {Abdeljaber}\ \emph {et~al.}(2017)\citenamefont {Abdeljaber}, \citenamefont {Avci}, \citenamefont {Kiranyaz}, \citenamefont {Gabbouj},\ and\ \citenamefont {Inman}}]{abdeljaber2017real}%
  \BibitemOpen
  \bibfield  {author} {\bibinfo {author} {\bibfnamefont {O.}~\bibnamefont {Abdeljaber}}, \bibinfo {author} {\bibfnamefont {O.}~\bibnamefont {Avci}}, \bibinfo {author} {\bibfnamefont {S.}~\bibnamefont {Kiranyaz}}, \bibinfo {author} {\bibfnamefont {M.}~\bibnamefont {Gabbouj}},\ and\ \bibinfo {author} {\bibfnamefont {D.~J.}\ \bibnamefont {Inman}},\ }\bibfield  {title} {\bibinfo {title} {Real-time vibration-based structural damage detection using one-dimensional convolutional neural networks},\ }\href@noop {} {\bibfield  {journal} {\bibinfo  {journal} {Journal of Sound and Vibration}\ }\textbf {\bibinfo {volume} {388}},\ \bibinfo {pages} {154} (\bibinfo {year} {2017})}\BibitemShut {NoStop}%
\bibitem [{\citenamefont {Ma}\ \emph {et~al.}(2022)\citenamefont {Ma}, \citenamefont {Fang}, \citenamefont {Wang}, \citenamefont {Xue}, \citenamefont {Dong},\ and\ \citenamefont {Wang}}]{ma2022real}%
  \BibitemOpen
  \bibfield  {author} {\bibinfo {author} {\bibfnamefont {D.}~\bibnamefont {Ma}}, \bibinfo {author} {\bibfnamefont {H.}~\bibnamefont {Fang}}, \bibinfo {author} {\bibfnamefont {N.}~\bibnamefont {Wang}}, \bibinfo {author} {\bibfnamefont {B.}~\bibnamefont {Xue}}, \bibinfo {author} {\bibfnamefont {J.}~\bibnamefont {Dong}},\ and\ \bibinfo {author} {\bibfnamefont {F.}~\bibnamefont {Wang}},\ }\bibfield  {title} {\bibinfo {title} {A real-time crack detection algorithm for pavement based on cnn with multiple feature layers},\ }\href@noop {} {\bibfield  {journal} {\bibinfo  {journal} {Road Materials and Pavement Design}\ }\textbf {\bibinfo {volume} {23}},\ \bibinfo {pages} {2115} (\bibinfo {year} {2022})}\BibitemShut {NoStop}%
\bibitem [{\citenamefont {Raissi}\ and\ \citenamefont {Karniadakis}(2018)}]{raissi2018hidden}%
  \BibitemOpen
  \bibfield  {author} {\bibinfo {author} {\bibfnamefont {M.}~\bibnamefont {Raissi}}\ and\ \bibinfo {author} {\bibfnamefont {G.~E.}\ \bibnamefont {Karniadakis}},\ }\bibfield  {title} {\bibinfo {title} {Hidden physics models: Machine learning of nonlinear partial differential equations},\ }\href@noop {} {\bibfield  {journal} {\bibinfo  {journal} {Journal of Computational Physics}\ }\textbf {\bibinfo {volume} {357}},\ \bibinfo {pages} {125} (\bibinfo {year} {2018})}\BibitemShut {NoStop}%
\bibitem [{\citenamefont {Raissi}\ \emph {et~al.}(2019)\citenamefont {Raissi}, \citenamefont {Perdikaris},\ and\ \citenamefont {Karniadakis}}]{raissi2019physics}%
  \BibitemOpen
  \bibfield  {author} {\bibinfo {author} {\bibfnamefont {M.}~\bibnamefont {Raissi}}, \bibinfo {author} {\bibfnamefont {P.}~\bibnamefont {Perdikaris}},\ and\ \bibinfo {author} {\bibfnamefont {G.~E.}\ \bibnamefont {Karniadakis}},\ }\bibfield  {title} {\bibinfo {title} {Physics-informed neural networks: A deep learning framework for solving forward and inverse problems involving nonlinear partial differential equations},\ }\href@noop {} {\bibfield  {journal} {\bibinfo  {journal} {Journal of Computational physics}\ }\textbf {\bibinfo {volume} {378}},\ \bibinfo {pages} {686} (\bibinfo {year} {2019})}\BibitemShut {NoStop}%
\bibitem [{\citenamefont {Bolandi}\ \emph {et~al.}(2023)\citenamefont {Bolandi}, \citenamefont {Sreekumar}, \citenamefont {Li}, \citenamefont {Lajnef},\ and\ \citenamefont {Boddeti}}]{bolandi2023physics}%
  \BibitemOpen
  \bibfield  {author} {\bibinfo {author} {\bibfnamefont {H.}~\bibnamefont {Bolandi}}, \bibinfo {author} {\bibfnamefont {G.}~\bibnamefont {Sreekumar}}, \bibinfo {author} {\bibfnamefont {X.}~\bibnamefont {Li}}, \bibinfo {author} {\bibfnamefont {N.}~\bibnamefont {Lajnef}},\ and\ \bibinfo {author} {\bibfnamefont {V.~N.}\ \bibnamefont {Boddeti}},\ }\bibfield  {title} {\bibinfo {title} {Physics informed neural network for dynamic stress prediction},\ }\href@noop {} {\bibfield  {journal} {\bibinfo  {journal} {Applied Intelligence}\ }\textbf {\bibinfo {volume} {53}},\ \bibinfo {pages} {26313} (\bibinfo {year} {2023})}\BibitemShut {NoStop}%
\bibitem [{\citenamefont {Parisi}\ \emph {et~al.}(2024)\citenamefont {Parisi}, \citenamefont {Ruggieri}, \citenamefont {Lovreglio}, \citenamefont {Fanti},\ and\ \citenamefont {Uva}}]{parisi2024use}%
  \BibitemOpen
  \bibfield  {author} {\bibinfo {author} {\bibfnamefont {F.}~\bibnamefont {Parisi}}, \bibinfo {author} {\bibfnamefont {S.}~\bibnamefont {Ruggieri}}, \bibinfo {author} {\bibfnamefont {R.}~\bibnamefont {Lovreglio}}, \bibinfo {author} {\bibfnamefont {M.~P.}\ \bibnamefont {Fanti}},\ and\ \bibinfo {author} {\bibfnamefont {G.}~\bibnamefont {Uva}},\ }\bibfield  {title} {\bibinfo {title} {On the use of mechanics-informed models to structural engineering systems: Application of graph neural networks for structural analysis},\ }in\ \href@noop {} {\emph {\bibinfo {booktitle} {Structures}}},\ Vol.~\bibinfo {volume} {59}\ (\bibinfo {year} {2024})\ p.\ \bibinfo {pages} {105712}\BibitemShut {NoStop}%
\bibitem [{\citenamefont {Song}\ \emph {et~al.}(2023)\citenamefont {Song}, \citenamefont {Wang}, \citenamefont {Fan},\ and\ \citenamefont {Lu}}]{song2023elastic}%
  \BibitemOpen
  \bibfield  {author} {\bibinfo {author} {\bibfnamefont {L.-H.}\ \bibnamefont {Song}}, \bibinfo {author} {\bibfnamefont {C.}~\bibnamefont {Wang}}, \bibinfo {author} {\bibfnamefont {J.-S.}\ \bibnamefont {Fan}},\ and\ \bibinfo {author} {\bibfnamefont {H.-M.}\ \bibnamefont {Lu}},\ }\bibfield  {title} {\bibinfo {title} {Elastic structural analysis based on graph neural network without labeled data},\ }\href@noop {} {\bibfield  {journal} {\bibinfo  {journal} {Computer-Aided Civil and Infrastructure Engineering}\ }\textbf {\bibinfo {volume} {38}},\ \bibinfo {pages} {1307} (\bibinfo {year} {2023})}\BibitemShut {NoStop}%
\bibitem [{\citenamefont {Chou}\ \emph {et~al.}(2024)\citenamefont {Chou}, \citenamefont {Chang}, \citenamefont {Jean}, \citenamefont {Chang}, \citenamefont {Huang},\ and\ \citenamefont {Chen}}]{chou2024structgnn}%
  \BibitemOpen
  \bibfield  {author} {\bibinfo {author} {\bibfnamefont {Y.-T.}\ \bibnamefont {Chou}}, \bibinfo {author} {\bibfnamefont {W.-T.}\ \bibnamefont {Chang}}, \bibinfo {author} {\bibfnamefont {J.~G.}\ \bibnamefont {Jean}}, \bibinfo {author} {\bibfnamefont {K.-H.}\ \bibnamefont {Chang}}, \bibinfo {author} {\bibfnamefont {Y.-N.}\ \bibnamefont {Huang}},\ and\ \bibinfo {author} {\bibfnamefont {C.-S.}\ \bibnamefont {Chen}},\ }\bibfield  {title} {\bibinfo {title} {Structgnn: An efficient graph neural network framework for static structural analysis},\ }\href@noop {} {\bibfield  {journal} {\bibinfo  {journal} {Computers \& Structures}\ }\textbf {\bibinfo {volume} {299}},\ \bibinfo {pages} {107385} (\bibinfo {year} {2024})}\BibitemShut {NoStop}%
\bibitem [{\citenamefont {Bloemheuvel}\ \emph {et~al.}(2021)\citenamefont {Bloemheuvel}, \citenamefont {van~den Hoogen},\ and\ \citenamefont {Atzmueller}}]{bloemheuvel2021computational}%
  \BibitemOpen
  \bibfield  {author} {\bibinfo {author} {\bibfnamefont {S.}~\bibnamefont {Bloemheuvel}}, \bibinfo {author} {\bibfnamefont {J.}~\bibnamefont {van~den Hoogen}},\ and\ \bibinfo {author} {\bibfnamefont {M.}~\bibnamefont {Atzmueller}},\ }\bibfield  {title} {\bibinfo {title} {A computational framework for modeling complex sensor network data using graph signal processing and graph neural networks in structural health monitoring},\ }\href@noop {} {\bibfield  {journal} {\bibinfo  {journal} {Applied Network Science}\ }\textbf {\bibinfo {volume} {6}},\ \bibinfo {pages} {97} (\bibinfo {year} {2021})}\BibitemShut {NoStop}%
\bibitem [{\citenamefont {Zhan}\ \emph {et~al.}(2023)\citenamefont {Zhan}, \citenamefont {Qin}, \citenamefont {Zhang},\ and\ \citenamefont {Sun}}]{zhan2023novel}%
  \BibitemOpen
  \bibfield  {author} {\bibinfo {author} {\bibfnamefont {P.}~\bibnamefont {Zhan}}, \bibinfo {author} {\bibfnamefont {X.}~\bibnamefont {Qin}}, \bibinfo {author} {\bibfnamefont {Q.}~\bibnamefont {Zhang}},\ and\ \bibinfo {author} {\bibfnamefont {Y.}~\bibnamefont {Sun}},\ }\bibfield  {title} {\bibinfo {title} {A novel structural damage detection method via multisensor spatial--temporal graph-based features and deep graph convolutional network},\ }\href@noop {} {\bibfield  {journal} {\bibinfo  {journal} {IEEE Transactions on Instrumentation and Measurement}\ }\textbf {\bibinfo {volume} {72}},\ \bibinfo {pages} {1} (\bibinfo {year} {2023})}\BibitemShut {NoStop}%
\bibitem [{\citenamefont {Hasni}\ \emph {et~al.}(2017{\natexlab{b}})\citenamefont {Hasni}, \citenamefont {Alavi}, \citenamefont {Jiao},\ and\ \citenamefont {Lajnef}}]{hasni2017detection}%
  \BibitemOpen
  \bibfield  {author} {\bibinfo {author} {\bibfnamefont {H.}~\bibnamefont {Hasni}}, \bibinfo {author} {\bibfnamefont {A.~H.}\ \bibnamefont {Alavi}}, \bibinfo {author} {\bibfnamefont {P.}~\bibnamefont {Jiao}},\ and\ \bibinfo {author} {\bibfnamefont {N.}~\bibnamefont {Lajnef}},\ }\bibfield  {title} {\bibinfo {title} {Detection of fatigue cracking in steel bridge girders: A support vector machine approach},\ }\href@noop {} {\bibfield  {journal} {\bibinfo  {journal} {Archives of Civil and Mechanical Engineering}\ }\textbf {\bibinfo {volume} {17}},\ \bibinfo {pages} {609} (\bibinfo {year} {2017}{\natexlab{b}})}\BibitemShut {NoStop}%
\bibitem [{\citenamefont {Gonz{\'a}lez}\ and\ \citenamefont {Zapico}(2008)}]{gonzalez2008seismic}%
  \BibitemOpen
  \bibfield  {author} {\bibinfo {author} {\bibfnamefont {M.~P.}\ \bibnamefont {Gonz{\'a}lez}}\ and\ \bibinfo {author} {\bibfnamefont {J.~L.}\ \bibnamefont {Zapico}},\ }\bibfield  {title} {\bibinfo {title} {Seismic damage identification in buildings using neural networks and modal data},\ }\href@noop {} {\bibfield  {journal} {\bibinfo  {journal} {Computers \& structures}\ }\textbf {\bibinfo {volume} {86}},\ \bibinfo {pages} {416} (\bibinfo {year} {2008})}\BibitemShut {NoStop}%
\bibitem [{\citenamefont {Rautela}\ and\ \citenamefont {Gopalakrishnan}(2021)}]{rautela2021ultrasonic}%
  \BibitemOpen
  \bibfield  {author} {\bibinfo {author} {\bibfnamefont {M.}~\bibnamefont {Rautela}}\ and\ \bibinfo {author} {\bibfnamefont {S.}~\bibnamefont {Gopalakrishnan}},\ }\bibfield  {title} {\bibinfo {title} {Ultrasonic guided wave based structural damage detection and localization using model assisted convolutional and recurrent neural networks},\ }\href@noop {} {\bibfield  {journal} {\bibinfo  {journal} {Expert Systems with Applications}\ }\textbf {\bibinfo {volume} {167}},\ \bibinfo {pages} {114189} (\bibinfo {year} {2021})}\BibitemShut {NoStop}%
\bibitem [{\citenamefont {Choe}\ \emph {et~al.}(2021)\citenamefont {Choe}, \citenamefont {Kim},\ and\ \citenamefont {Kim}}]{choe2021sequence}%
  \BibitemOpen
  \bibfield  {author} {\bibinfo {author} {\bibfnamefont {D.-E.}\ \bibnamefont {Choe}}, \bibinfo {author} {\bibfnamefont {H.-C.}\ \bibnamefont {Kim}},\ and\ \bibinfo {author} {\bibfnamefont {M.-H.}\ \bibnamefont {Kim}},\ }\bibfield  {title} {\bibinfo {title} {Sequence-based modeling of deep learning with {LSTM} and {GRU} networks for structural damage detection of floating offshore wind turbine blades},\ }\href@noop {} {\bibfield  {journal} {\bibinfo  {journal} {Renewable Energy}\ }\textbf {\bibinfo {volume} {174}},\ \bibinfo {pages} {218} (\bibinfo {year} {2021})}\BibitemShut {NoStop}%
\bibitem [{\citenamefont {Xu}\ \emph {et~al.}(2019)\citenamefont {Xu}, \citenamefont {Lin}, \citenamefont {Zhan},\ and\ \citenamefont {Wang}}]{xu2019multistage}%
  \BibitemOpen
  \bibfield  {author} {\bibinfo {author} {\bibfnamefont {Y.-L.}\ \bibnamefont {Xu}}, \bibinfo {author} {\bibfnamefont {J.-F.}\ \bibnamefont {Lin}}, \bibinfo {author} {\bibfnamefont {S.}~\bibnamefont {Zhan}},\ and\ \bibinfo {author} {\bibfnamefont {F.-Y.}\ \bibnamefont {Wang}},\ }\bibfield  {title} {\bibinfo {title} {Multistage damage detection of a transmission tower: Numerical investigation and experimental validation},\ }\href@noop {} {\bibfield  {journal} {\bibinfo  {journal} {Structural Control and Health Monitoring}\ }\textbf {\bibinfo {volume} {26}},\ \bibinfo {pages} {e2366} (\bibinfo {year} {2019})}\BibitemShut {NoStop}%
\bibitem [{\citenamefont {Gulgec}\ \emph {et~al.}(2019)\citenamefont {Gulgec}, \citenamefont {Tak{\'a}{\v{c}}},\ and\ \citenamefont {Pakzad}}]{gulgec2019convolutional}%
  \BibitemOpen
  \bibfield  {author} {\bibinfo {author} {\bibfnamefont {N.~S.}\ \bibnamefont {Gulgec}}, \bibinfo {author} {\bibfnamefont {M.}~\bibnamefont {Tak{\'a}{\v{c}}}},\ and\ \bibinfo {author} {\bibfnamefont {S.~N.}\ \bibnamefont {Pakzad}},\ }\bibfield  {title} {\bibinfo {title} {Convolutional neural network approach for robust structural damage detection and localization},\ }\href@noop {} {\bibfield  {journal} {\bibinfo  {journal} {Journal of computing in civil engineering}\ }\textbf {\bibinfo {volume} {33}},\ \bibinfo {pages} {04019005} (\bibinfo {year} {2019})}\BibitemShut {NoStop}%
\bibitem [{\citenamefont {Mustafa}\ \emph {et~al.}(2023)\citenamefont {Mustafa}, \citenamefont {Sekiya},\ and\ \citenamefont {Hirano}}]{mustafa2023evaluation}%
  \BibitemOpen
  \bibfield  {author} {\bibinfo {author} {\bibfnamefont {S.}~\bibnamefont {Mustafa}}, \bibinfo {author} {\bibfnamefont {H.}~\bibnamefont {Sekiya}},\ and\ \bibinfo {author} {\bibfnamefont {S.}~\bibnamefont {Hirano}},\ }\bibfield  {title} {\bibinfo {title} {Evaluation of fatigue damage in steel girder bridges using displacement influence lines},\ }in\ \href@noop {} {\emph {\bibinfo {booktitle} {Structures}}},\ Vol.~\bibinfo {volume} {53}\ (\bibinfo {organization} {Elsevier},\ \bibinfo {year} {2023})\ pp.\ \bibinfo {pages} {1160--1171}\BibitemShut {NoStop}%
\bibitem [{\citenamefont {Li}\ \emph {et~al.}(2022)\citenamefont {Li}, \citenamefont {Salem}, \citenamefont {Bolandi}, \citenamefont {Boddeti},\ and\ \citenamefont {Lajnef}}]{li2022methods}%
  \BibitemOpen
  \bibfield  {author} {\bibinfo {author} {\bibfnamefont {X.}~\bibnamefont {Li}}, \bibinfo {author} {\bibfnamefont {T.}~\bibnamefont {Salem}}, \bibinfo {author} {\bibfnamefont {H.}~\bibnamefont {Bolandi}}, \bibinfo {author} {\bibfnamefont {V.}~\bibnamefont {Boddeti}},\ and\ \bibinfo {author} {\bibfnamefont {N.}~\bibnamefont {Lajnef}},\ }\bibfield  {title} {\bibinfo {title} {Methods for the rapid detection of boundary condition variations in structural systems},\ }in\ \href@noop {} {\emph {\bibinfo {booktitle} {Smart Materials, Adaptive Structures and Intelligent Systems}}},\ Vol.\ \bibinfo {volume} {86274}\ (\bibinfo {year} {2022})\ p.\ \bibinfo {pages} {V001T05A004}\BibitemShut {NoStop}%
\bibitem [{\citenamefont {Wu}\ \emph {et~al.}(2018)\citenamefont {Wu}, \citenamefont {Wu}, \citenamefont {Yang},\ and\ \citenamefont {He}}]{wu2018damage}%
  \BibitemOpen
  \bibfield  {author} {\bibinfo {author} {\bibfnamefont {B.}~\bibnamefont {Wu}}, \bibinfo {author} {\bibfnamefont {G.}~\bibnamefont {Wu}}, \bibinfo {author} {\bibfnamefont {C.}~\bibnamefont {Yang}},\ and\ \bibinfo {author} {\bibfnamefont {Y.}~\bibnamefont {He}},\ }\bibfield  {title} {\bibinfo {title} {Damage identification method for continuous girder bridges based on spatially-distributed long-gauge strain sensing under moving loads},\ }\href@noop {} {\bibfield  {journal} {\bibinfo  {journal} {Mechanical Systems and Signal Processing}\ }\textbf {\bibinfo {volume} {104}},\ \bibinfo {pages} {415} (\bibinfo {year} {2018})}\BibitemShut {NoStop}%
\bibitem [{\citenamefont {Li}\ \emph {et~al.}(2021)\citenamefont {Li}, \citenamefont {Wang}, \citenamefont {Liu},\ and\ \citenamefont {Lin}}]{li2021deep}%
  \BibitemOpen
  \bibfield  {author} {\bibinfo {author} {\bibfnamefont {T.}~\bibnamefont {Li}}, \bibinfo {author} {\bibfnamefont {Z.}~\bibnamefont {Wang}}, \bibinfo {author} {\bibfnamefont {S.}~\bibnamefont {Liu}},\ and\ \bibinfo {author} {\bibfnamefont {W.-Y.}\ \bibnamefont {Lin}},\ }\bibfield  {title} {\bibinfo {title} {Deep unsupervised anomaly detection},\ }in\ \href@noop {} {\emph {\bibinfo {booktitle} {Proceedings of the IEEE/CVF Winter Conference on Applications of Computer Vision}}}\ (\bibinfo {year} {2021})\ pp.\ \bibinfo {pages} {3636--3645}\BibitemShut {NoStop}%
\bibitem [{\citenamefont {Heo}\ \emph {et~al.}(2018)\citenamefont {Heo}, \citenamefont {Kim}, \citenamefont {Jeon},\ and\ \citenamefont {Jeon}}]{heo2018experimental}%
  \BibitemOpen
  \bibfield  {author} {\bibinfo {author} {\bibfnamefont {G.}~\bibnamefont {Heo}}, \bibinfo {author} {\bibfnamefont {C.}~\bibnamefont {Kim}}, \bibinfo {author} {\bibfnamefont {S.}~\bibnamefont {Jeon}},\ and\ \bibinfo {author} {\bibfnamefont {J.}~\bibnamefont {Jeon}},\ }\bibfield  {title} {\bibinfo {title} {An experimental study of a data compression technology-based intelligent data acquisition ({IDAQ}) system for structural health monitoring of a long-span bridge},\ }\href@noop {} {\bibfield  {journal} {\bibinfo  {journal} {Applied Sciences}\ }\textbf {\bibinfo {volume} {8}},\ \bibinfo {pages} {361} (\bibinfo {year} {2018})}\BibitemShut {NoStop}%
\bibitem [{\citenamefont {Chen}\ \emph {et~al.}(2019)\citenamefont {Chen}, \citenamefont {Wu},\ and\ \citenamefont {Feng}}]{chen2019damage}%
  \BibitemOpen
  \bibfield  {author} {\bibinfo {author} {\bibfnamefont {S.-Z.}\ \bibnamefont {Chen}}, \bibinfo {author} {\bibfnamefont {G.}~\bibnamefont {Wu}},\ and\ \bibinfo {author} {\bibfnamefont {D.-C.}\ \bibnamefont {Feng}},\ }\bibfield  {title} {\bibinfo {title} {Damage detection of highway bridges based on long-gauge strain response under stochastic traffic flow},\ }\href@noop {} {\bibfield  {journal} {\bibinfo  {journal} {Mechanical Systems and Signal Processing}\ }\textbf {\bibinfo {volume} {127}},\ \bibinfo {pages} {551} (\bibinfo {year} {2019})}\BibitemShut {NoStop}%
\bibitem [{\citenamefont {Azim}\ and\ \citenamefont {G{\"u}l}(2021{\natexlab{a}})}]{azim2021data}%
  \BibitemOpen
  \bibfield  {author} {\bibinfo {author} {\bibfnamefont {M.~R.}\ \bibnamefont {Azim}}\ and\ \bibinfo {author} {\bibfnamefont {M.}~\bibnamefont {G{\"u}l}},\ }\bibfield  {title} {\bibinfo {title} {Data-driven damage identification technique for steel truss railroad bridges utilizing principal component analysis of strain response},\ }\href@noop {} {\bibfield  {journal} {\bibinfo  {journal} {Structure and Infrastructure Engineering}\ }\textbf {\bibinfo {volume} {17}},\ \bibinfo {pages} {1019} (\bibinfo {year} {2021}{\natexlab{a}})}\BibitemShut {NoStop}%
\bibitem [{\citenamefont {Azim}\ and\ \citenamefont {G{\"u}l}(2021{\natexlab{b}})}]{azim2021development}%
  \BibitemOpen
  \bibfield  {author} {\bibinfo {author} {\bibfnamefont {M.~R.}\ \bibnamefont {Azim}}\ and\ \bibinfo {author} {\bibfnamefont {M.}~\bibnamefont {G{\"u}l}},\ }\bibfield  {title} {\bibinfo {title} {Development of a novel damage detection framework for truss railway bridges using operational acceleration and strain response},\ }\href@noop {} {\bibfield  {journal} {\bibinfo  {journal} {Vibration}\ }\textbf {\bibinfo {volume} {4}},\ \bibinfo {pages} {422} (\bibinfo {year} {2021}{\natexlab{b}})}\BibitemShut {NoStop}%
\bibitem [{\citenamefont {Rastin}\ \emph {et~al.}(2021)\citenamefont {Rastin}, \citenamefont {Ghodrati~Amiri},\ and\ \citenamefont {Darvishan}}]{rastin2021unsupervised}%
  \BibitemOpen
  \bibfield  {author} {\bibinfo {author} {\bibfnamefont {Z.}~\bibnamefont {Rastin}}, \bibinfo {author} {\bibfnamefont {G.}~\bibnamefont {Ghodrati~Amiri}},\ and\ \bibinfo {author} {\bibfnamefont {E.}~\bibnamefont {Darvishan}},\ }\bibfield  {title} {\bibinfo {title} {Unsupervised structural damage detection technique based on a deep convolutional autoencoder},\ }\href@noop {} {\bibfield  {journal} {\bibinfo  {journal} {Shock and Vibration}\ }\textbf {\bibinfo {volume} {2021}},\ \bibinfo {pages} {1} (\bibinfo {year} {2021})}\BibitemShut {NoStop}%
\bibitem [{\citenamefont {Ni}\ \emph {et~al.}(2020)\citenamefont {Ni}, \citenamefont {Zhang},\ and\ \citenamefont {Noori}}]{ni2020deep}%
  \BibitemOpen
  \bibfield  {author} {\bibinfo {author} {\bibfnamefont {F.}~\bibnamefont {Ni}}, \bibinfo {author} {\bibfnamefont {J.}~\bibnamefont {Zhang}},\ and\ \bibinfo {author} {\bibfnamefont {M.~N.}\ \bibnamefont {Noori}},\ }\bibfield  {title} {\bibinfo {title} {Deep learning for data anomaly detection and data compression of a long-span suspension bridge},\ }\href@noop {} {\bibfield  {journal} {\bibinfo  {journal} {Computer-Aided Civil and Infrastructure Engineering}\ }\textbf {\bibinfo {volume} {35}},\ \bibinfo {pages} {685} (\bibinfo {year} {2020})}\BibitemShut {NoStop}%
\bibitem [{\citenamefont {Giglioni}\ \emph {et~al.}(2023)\citenamefont {Giglioni}, \citenamefont {Venanzi}, \citenamefont {Poggioni}, \citenamefont {Milani},\ and\ \citenamefont {Ubertini}}]{giglioni2023autoencoders}%
  \BibitemOpen
  \bibfield  {author} {\bibinfo {author} {\bibfnamefont {V.}~\bibnamefont {Giglioni}}, \bibinfo {author} {\bibfnamefont {I.}~\bibnamefont {Venanzi}}, \bibinfo {author} {\bibfnamefont {V.}~\bibnamefont {Poggioni}}, \bibinfo {author} {\bibfnamefont {A.}~\bibnamefont {Milani}},\ and\ \bibinfo {author} {\bibfnamefont {F.}~\bibnamefont {Ubertini}},\ }\bibfield  {title} {\bibinfo {title} {Autoencoders for unsupervised real-time bridge health assessment},\ }\href@noop {} {\bibfield  {journal} {\bibinfo  {journal} {Computer-Aided Civil and Infrastructure Engineering}\ }\textbf {\bibinfo {volume} {38}},\ \bibinfo {pages} {959} (\bibinfo {year} {2023})}\BibitemShut {NoStop}%
\bibitem [{\citenamefont {Khoshnoudian}\ \emph {et~al.}(2017)\citenamefont {Khoshnoudian}, \citenamefont {Talaei},\ and\ \citenamefont {Fallahian}}]{khoshnoudian2017structural}%
  \BibitemOpen
  \bibfield  {author} {\bibinfo {author} {\bibfnamefont {F.}~\bibnamefont {Khoshnoudian}}, \bibinfo {author} {\bibfnamefont {S.}~\bibnamefont {Talaei}},\ and\ \bibinfo {author} {\bibfnamefont {M.}~\bibnamefont {Fallahian}},\ }\bibfield  {title} {\bibinfo {title} {Structural damage detection using {FRF} data, {2D-PCA}, artificial neural networks and imperialist competitive algorithm simultaneously},\ }\href@noop {} {\bibfield  {journal} {\bibinfo  {journal} {International Journal of Structural Stability and Dynamics}\ }\textbf {\bibinfo {volume} {17}},\ \bibinfo {pages} {1750073} (\bibinfo {year} {2017})}\BibitemShut {NoStop}%
\bibitem [{\citenamefont {Cao}\ \emph {et~al.}(2019)\citenamefont {Cao}, \citenamefont {Ouyang},\ and\ \citenamefont {Cheng}}]{cao2019baseline}%
  \BibitemOpen
  \bibfield  {author} {\bibinfo {author} {\bibfnamefont {S.}~\bibnamefont {Cao}}, \bibinfo {author} {\bibfnamefont {H.}~\bibnamefont {Ouyang}},\ and\ \bibinfo {author} {\bibfnamefont {L.}~\bibnamefont {Cheng}},\ }\bibfield  {title} {\bibinfo {title} {Baseline-free adaptive damage localization of plate-type structures by using robust pca and gaussian smoothing},\ }\href@noop {} {\bibfield  {journal} {\bibinfo  {journal} {Mechanical Systems and Signal Processing}\ }\textbf {\bibinfo {volume} {122}},\ \bibinfo {pages} {232} (\bibinfo {year} {2019})}\BibitemShut {NoStop}%
\bibitem [{\citenamefont {Sen}\ \emph {et~al.}(2019)\citenamefont {Sen}, \citenamefont {Erazo}, \citenamefont {Zhang}, \citenamefont {Nagarajaiah},\ and\ \citenamefont {Sun}}]{sen2019effectiveness}%
  \BibitemOpen
  \bibfield  {author} {\bibinfo {author} {\bibfnamefont {D.}~\bibnamefont {Sen}}, \bibinfo {author} {\bibfnamefont {K.}~\bibnamefont {Erazo}}, \bibinfo {author} {\bibfnamefont {W.}~\bibnamefont {Zhang}}, \bibinfo {author} {\bibfnamefont {S.}~\bibnamefont {Nagarajaiah}},\ and\ \bibinfo {author} {\bibfnamefont {L.}~\bibnamefont {Sun}},\ }\bibfield  {title} {\bibinfo {title} {On the effectiveness of principal component analysis for decoupling structural damage and environmental effects in bridge structures},\ }\href@noop {} {\bibfield  {journal} {\bibinfo  {journal} {Journal of Sound and Vibration}\ }\textbf {\bibinfo {volume} {457}},\ \bibinfo {pages} {280} (\bibinfo {year} {2019})}\BibitemShut {NoStop}%
\bibitem [{\citenamefont {Zhang}\ and\ \citenamefont {Sun}(2021)}]{zhang2021structural}%
  \BibitemOpen
  \bibfield  {author} {\bibinfo {author} {\bibfnamefont {Z.}~\bibnamefont {Zhang}}\ and\ \bibinfo {author} {\bibfnamefont {C.}~\bibnamefont {Sun}},\ }\bibfield  {title} {\bibinfo {title} {Structural damage identification via physics-guided machine learning: a methodology integrating pattern recognition with finite element model updating},\ }\href@noop {} {\bibfield  {journal} {\bibinfo  {journal} {Structural Health Monitoring}\ }\textbf {\bibinfo {volume} {20}},\ \bibinfo {pages} {1675} (\bibinfo {year} {2021})}\BibitemShut {NoStop}%
\bibitem [{\citenamefont {Zhang}\ \emph {et~al.}(2021)\citenamefont {Zhang}, \citenamefont {Li},\ and\ \citenamefont {Ye}}]{zhang2021damage}%
  \BibitemOpen
  \bibfield  {author} {\bibinfo {author} {\bibfnamefont {S.}~\bibnamefont {Zhang}}, \bibinfo {author} {\bibfnamefont {C.~M.}\ \bibnamefont {Li}},\ and\ \bibinfo {author} {\bibfnamefont {W.}~\bibnamefont {Ye}},\ }\bibfield  {title} {\bibinfo {title} {Damage localization in plate-like structures using time-varying feature and one-dimensional convolutional neural network},\ }\href@noop {} {\bibfield  {journal} {\bibinfo  {journal} {Mechanical Systems and Signal Processing}\ }\textbf {\bibinfo {volume} {147}},\ \bibinfo {pages} {107107} (\bibinfo {year} {2021})}\BibitemShut {NoStop}%
\bibitem [{\citenamefont {Cofre-Martel}\ \emph {et~al.}(2019)\citenamefont {Cofre-Martel}, \citenamefont {Kobrich}, \citenamefont {Lopez~Droguett},\ and\ \citenamefont {Meruane}}]{cofre2019deep}%
  \BibitemOpen
  \bibfield  {author} {\bibinfo {author} {\bibfnamefont {S.}~\bibnamefont {Cofre-Martel}}, \bibinfo {author} {\bibfnamefont {P.}~\bibnamefont {Kobrich}}, \bibinfo {author} {\bibfnamefont {E.}~\bibnamefont {Lopez~Droguett}},\ and\ \bibinfo {author} {\bibfnamefont {V.}~\bibnamefont {Meruane}},\ }\bibfield  {title} {\bibinfo {title} {Deep convolutional neural network-based structural damage localization and quantification using transmissibility data},\ }\href@noop {} {\bibfield  {journal} {\bibinfo  {journal} {Shock and Vibration}\ }\textbf {\bibinfo {volume} {2019}} (\bibinfo {year} {2019})}\BibitemShut {NoStop}%
\bibitem [{\citenamefont {Alavi}\ \emph {et~al.}(2016{\natexlab{a}})\citenamefont {Alavi}, \citenamefont {Hasni}, \citenamefont {Lajnef}, \citenamefont {Chatti},\ and\ \citenamefont {Faridazar}}]{alavi2016intelligent}%
  \BibitemOpen
  \bibfield  {author} {\bibinfo {author} {\bibfnamefont {A.~H.}\ \bibnamefont {Alavi}}, \bibinfo {author} {\bibfnamefont {H.}~\bibnamefont {Hasni}}, \bibinfo {author} {\bibfnamefont {N.}~\bibnamefont {Lajnef}}, \bibinfo {author} {\bibfnamefont {K.}~\bibnamefont {Chatti}},\ and\ \bibinfo {author} {\bibfnamefont {F.}~\bibnamefont {Faridazar}},\ }\bibfield  {title} {\bibinfo {title} {An intelligent structural damage detection approach based on self-powered wireless sensor data},\ }\href@noop {} {\bibfield  {journal} {\bibinfo  {journal} {Automation in Construction}\ }\textbf {\bibinfo {volume} {62}},\ \bibinfo {pages} {24} (\bibinfo {year} {2016}{\natexlab{a}})}\BibitemShut {NoStop}%
\bibitem [{\citenamefont {Alavi}\ \emph {et~al.}(2016{\natexlab{b}})\citenamefont {Alavi}, \citenamefont {Hasni}, \citenamefont {Lajnef}, \citenamefont {Chatti},\ and\ \citenamefont {Faridazar}}]{alavi2016damage}%
  \BibitemOpen
  \bibfield  {author} {\bibinfo {author} {\bibfnamefont {A.~H.}\ \bibnamefont {Alavi}}, \bibinfo {author} {\bibfnamefont {H.}~\bibnamefont {Hasni}}, \bibinfo {author} {\bibfnamefont {N.}~\bibnamefont {Lajnef}}, \bibinfo {author} {\bibfnamefont {K.}~\bibnamefont {Chatti}},\ and\ \bibinfo {author} {\bibfnamefont {F.}~\bibnamefont {Faridazar}},\ }\bibfield  {title} {\bibinfo {title} {Damage detection using self-powered wireless sensor data: An evolutionary approach},\ }\href@noop {} {\bibfield  {journal} {\bibinfo  {journal} {Measurement}\ }\textbf {\bibinfo {volume} {82}},\ \bibinfo {pages} {254} (\bibinfo {year} {2016}{\natexlab{b}})}\BibitemShut {NoStop}%
\bibitem [{\citenamefont {Hasni}\ \emph {et~al.}(2017{\natexlab{c}})\citenamefont {Hasni}, \citenamefont {Alavi}, \citenamefont {Chatti},\ and\ \citenamefont {Lajnef}}]{hasni2017continuous}%
  \BibitemOpen
  \bibfield  {author} {\bibinfo {author} {\bibfnamefont {H.}~\bibnamefont {Hasni}}, \bibinfo {author} {\bibfnamefont {A.}~\bibnamefont {Alavi}}, \bibinfo {author} {\bibfnamefont {K.}~\bibnamefont {Chatti}},\ and\ \bibinfo {author} {\bibfnamefont {N.}~\bibnamefont {Lajnef}},\ }\bibfield  {title} {\bibinfo {title} {Continuous health monitoring of asphalt concrete pavements using surface-mounted battery-free wireless sensors},\ }in\ \href@noop {} {\emph {\bibinfo {booktitle} {Bearing Capacity of Roads, Railways and Airfields}}}\ (\bibinfo  {publisher} {CRC Press},\ \bibinfo {year} {2017})\ pp.\ \bibinfo {pages} {637--643}\BibitemShut {NoStop}%
\bibitem [{\citenamefont {Salehi}\ \emph {et~al.}(2021)\citenamefont {Salehi}, \citenamefont {Burgue{\~n}o}, \citenamefont {Chakrabartty}, \citenamefont {Lajnef},\ and\ \citenamefont {Alavi}}]{salehi2021comprehensive}%
  \BibitemOpen
  \bibfield  {author} {\bibinfo {author} {\bibfnamefont {H.}~\bibnamefont {Salehi}}, \bibinfo {author} {\bibfnamefont {R.}~\bibnamefont {Burgue{\~n}o}}, \bibinfo {author} {\bibfnamefont {S.}~\bibnamefont {Chakrabartty}}, \bibinfo {author} {\bibfnamefont {N.}~\bibnamefont {Lajnef}},\ and\ \bibinfo {author} {\bibfnamefont {A.~H.}\ \bibnamefont {Alavi}},\ }\bibfield  {title} {\bibinfo {title} {A comprehensive review of self-powered sensors in civil infrastructure: State-of-the-art and future research trends},\ }\href@noop {} {\bibfield  {journal} {\bibinfo  {journal} {Engineering Structures}\ }\textbf {\bibinfo {volume} {234}},\ \bibinfo {pages} {111963} (\bibinfo {year} {2021})}\BibitemShut {NoStop}%
\bibitem [{\citenamefont {Bewick}\ \emph {et~al.}(2004)\citenamefont {Bewick}, \citenamefont {Cheek},\ and\ \citenamefont {Ball}}]{bewick2004statistics}%
  \BibitemOpen
  \bibfield  {author} {\bibinfo {author} {\bibfnamefont {V.}~\bibnamefont {Bewick}}, \bibinfo {author} {\bibfnamefont {L.}~\bibnamefont {Cheek}},\ and\ \bibinfo {author} {\bibfnamefont {J.}~\bibnamefont {Ball}},\ }\bibfield  {title} {\bibinfo {title} {Statistics review 13: receiver operating characteristic curves},\ }\href@noop {} {\bibfield  {journal} {\bibinfo  {journal} {Critical Care}\ }\textbf {\bibinfo {volume} {8}},\ \bibinfo {pages} {1} (\bibinfo {year} {2004})}\BibitemShut {NoStop}%
\bibitem [{\citenamefont {Liu}\ \emph {et~al.}(2008)\citenamefont {Liu}, \citenamefont {Ting},\ and\ \citenamefont {Zhou}}]{liu2008isolation}%
  \BibitemOpen
  \bibfield  {author} {\bibinfo {author} {\bibfnamefont {F.~T.}\ \bibnamefont {Liu}}, \bibinfo {author} {\bibfnamefont {K.~M.}\ \bibnamefont {Ting}},\ and\ \bibinfo {author} {\bibfnamefont {Z.-H.}\ \bibnamefont {Zhou}},\ }\bibfield  {title} {\bibinfo {title} {Isolation forest},\ }in\ \href@noop {} {\emph {\bibinfo {booktitle} {IEEE International Conference on Data Mining}}}\ (\bibinfo {year} {2008})\ pp.\ \bibinfo {pages} {413--422}\BibitemShut {NoStop}%
\bibitem [{\citenamefont {Pevn{\`y}}(2016)}]{pevny2016loda}%
  \BibitemOpen
  \bibfield  {author} {\bibinfo {author} {\bibfnamefont {T.}~\bibnamefont {Pevn{\`y}}},\ }\bibfield  {title} {\bibinfo {title} {Loda: Lightweight on-line detector of anomalies},\ }\href@noop {} {\bibfield  {journal} {\bibinfo  {journal} {Machine Learning}\ }\textbf {\bibinfo {volume} {102}},\ \bibinfo {pages} {275} (\bibinfo {year} {2016})}\BibitemShut {NoStop}%
\bibitem [{\citenamefont {Papadimitriou}\ \emph {et~al.}(2005)\citenamefont {Papadimitriou}, \citenamefont {Sun},\ and\ \citenamefont {Faloutsos}}]{papadimitriou2005streaming}%
  \BibitemOpen
  \bibfield  {author} {\bibinfo {author} {\bibfnamefont {S.}~\bibnamefont {Papadimitriou}}, \bibinfo {author} {\bibfnamefont {J.}~\bibnamefont {Sun}},\ and\ \bibinfo {author} {\bibfnamefont {C.}~\bibnamefont {Faloutsos}},\ }\bibfield  {title} {\bibinfo {title} {Streaming pattern discovery in multiple time-series}\ }(\bibinfo  {publisher} {Carnegie Mellon University},\ \bibinfo {year} {2005})\BibitemShut {NoStop}%
\bibitem [{\citenamefont {Shafer}\ \emph {et~al.}(2012)\citenamefont {Shafer}, \citenamefont {Ren}, \citenamefont {Boddeti}, \citenamefont {Abe}, \citenamefont {Ganger},\ and\ \citenamefont {Faloutsos}}]{shafer2012rainmon}%
  \BibitemOpen
  \bibfield  {author} {\bibinfo {author} {\bibfnamefont {I.}~\bibnamefont {Shafer}}, \bibinfo {author} {\bibfnamefont {K.}~\bibnamefont {Ren}}, \bibinfo {author} {\bibfnamefont {V.~N.}\ \bibnamefont {Boddeti}}, \bibinfo {author} {\bibfnamefont {Y.}~\bibnamefont {Abe}}, \bibinfo {author} {\bibfnamefont {G.~R.}\ \bibnamefont {Ganger}},\ and\ \bibinfo {author} {\bibfnamefont {C.}~\bibnamefont {Faloutsos}},\ }\bibfield  {title} {\bibinfo {title} {Rainmon: An integrated approach to mining bursty timeseries monitoring data},\ }in\ \href@noop {} {\emph {\bibinfo {booktitle} {Proceedings of the ACM SIGKDD International Conference on Knowledge Discovery and Data Mining}}}\ (\bibinfo {year} {2012})\ pp.\ \bibinfo {pages} {1158--1166}\BibitemShut {NoStop}%
\bibitem [{\citenamefont {Bolandi}\ \emph {et~al.}(2019)\citenamefont {Bolandi}, \citenamefont {Lajnef}, \citenamefont {Jiao}, \citenamefont {Barri}, \citenamefont {Hasni},\ and\ \citenamefont {Alavi}}]{bolandi2019novel}%
  \BibitemOpen
  \bibfield  {author} {\bibinfo {author} {\bibfnamefont {H.}~\bibnamefont {Bolandi}}, \bibinfo {author} {\bibfnamefont {N.}~\bibnamefont {Lajnef}}, \bibinfo {author} {\bibfnamefont {P.}~\bibnamefont {Jiao}}, \bibinfo {author} {\bibfnamefont {K.}~\bibnamefont {Barri}}, \bibinfo {author} {\bibfnamefont {H.}~\bibnamefont {Hasni}},\ and\ \bibinfo {author} {\bibfnamefont {A.~H.}\ \bibnamefont {Alavi}},\ }\bibfield  {title} {\bibinfo {title} {A novel data reduction approach for structural health monitoring systems},\ }\href@noop {} {\bibfield  {journal} {\bibinfo  {journal} {Sensors}\ }\textbf {\bibinfo {volume} {19}},\ \bibinfo {pages} {4823} (\bibinfo {year} {2019})}\BibitemShut {NoStop}%
\bibitem [{\citenamefont {Hasni}\ \emph {et~al.}(2018)\citenamefont {Hasni}, \citenamefont {Jiao}, \citenamefont {Lajnef},\ and\ \citenamefont {Alavi}}]{hasni2018damage}%
  \BibitemOpen
  \bibfield  {author} {\bibinfo {author} {\bibfnamefont {H.}~\bibnamefont {Hasni}}, \bibinfo {author} {\bibfnamefont {P.}~\bibnamefont {Jiao}}, \bibinfo {author} {\bibfnamefont {N.}~\bibnamefont {Lajnef}},\ and\ \bibinfo {author} {\bibfnamefont {A.~H.}\ \bibnamefont {Alavi}},\ }\bibfield  {title} {\bibinfo {title} {Damage localization and quantification in gusset plates: A battery-free sensing approach},\ }\href@noop {} {\bibfield  {journal} {\bibinfo  {journal} {Structural Control and Health Monitoring}\ }\textbf {\bibinfo {volume} {25}},\ \bibinfo {pages} {e2158} (\bibinfo {year} {2018})}\BibitemShut {NoStop}%
\bibitem [{\citenamefont {Chawla}\ \emph {et~al.}(2002)\citenamefont {Chawla}, \citenamefont {Bowyer}, \citenamefont {Hall},\ and\ \citenamefont {Kegelmeyer}}]{chawla2002smote}%
  \BibitemOpen
  \bibfield  {author} {\bibinfo {author} {\bibfnamefont {N.~V.}\ \bibnamefont {Chawla}}, \bibinfo {author} {\bibfnamefont {K.~W.}\ \bibnamefont {Bowyer}}, \bibinfo {author} {\bibfnamefont {L.~O.}\ \bibnamefont {Hall}},\ and\ \bibinfo {author} {\bibfnamefont {W.~P.}\ \bibnamefont {Kegelmeyer}},\ }\bibfield  {title} {\bibinfo {title} {Smote: synthetic minority over-sampling technique},\ }\href@noop {} {\bibfield  {journal} {\bibinfo  {journal} {Journal of artificial intelligence research}\ }\textbf {\bibinfo {volume} {16}},\ \bibinfo {pages} {321} (\bibinfo {year} {2002})}\BibitemShut {NoStop}%
\bibitem [{\citenamefont {Wilson}(1972)}]{wilson1972asymptotic}%
  \BibitemOpen
  \bibfield  {author} {\bibinfo {author} {\bibfnamefont {D.~L.}\ \bibnamefont {Wilson}},\ }\bibfield  {title} {\bibinfo {title} {Asymptotic properties of nearest neighbor rules using edited data},\ }\href@noop {} {\bibfield  {journal} {\bibinfo  {journal} {IEEE Transactions on Systems, Man, and Cybernetics}\ ,\ \bibinfo {pages} {408}} (\bibinfo {year} {1972})}\BibitemShut {NoStop}%
\end{thebibliography}
% \includepdf[pages={1,2,3}]{Supplementary.pdf}
\clearpage
\onecolumngrid
\includegraphics[page={1},scale=0.96,trim=2cm 0cm 0 2cm,clip]{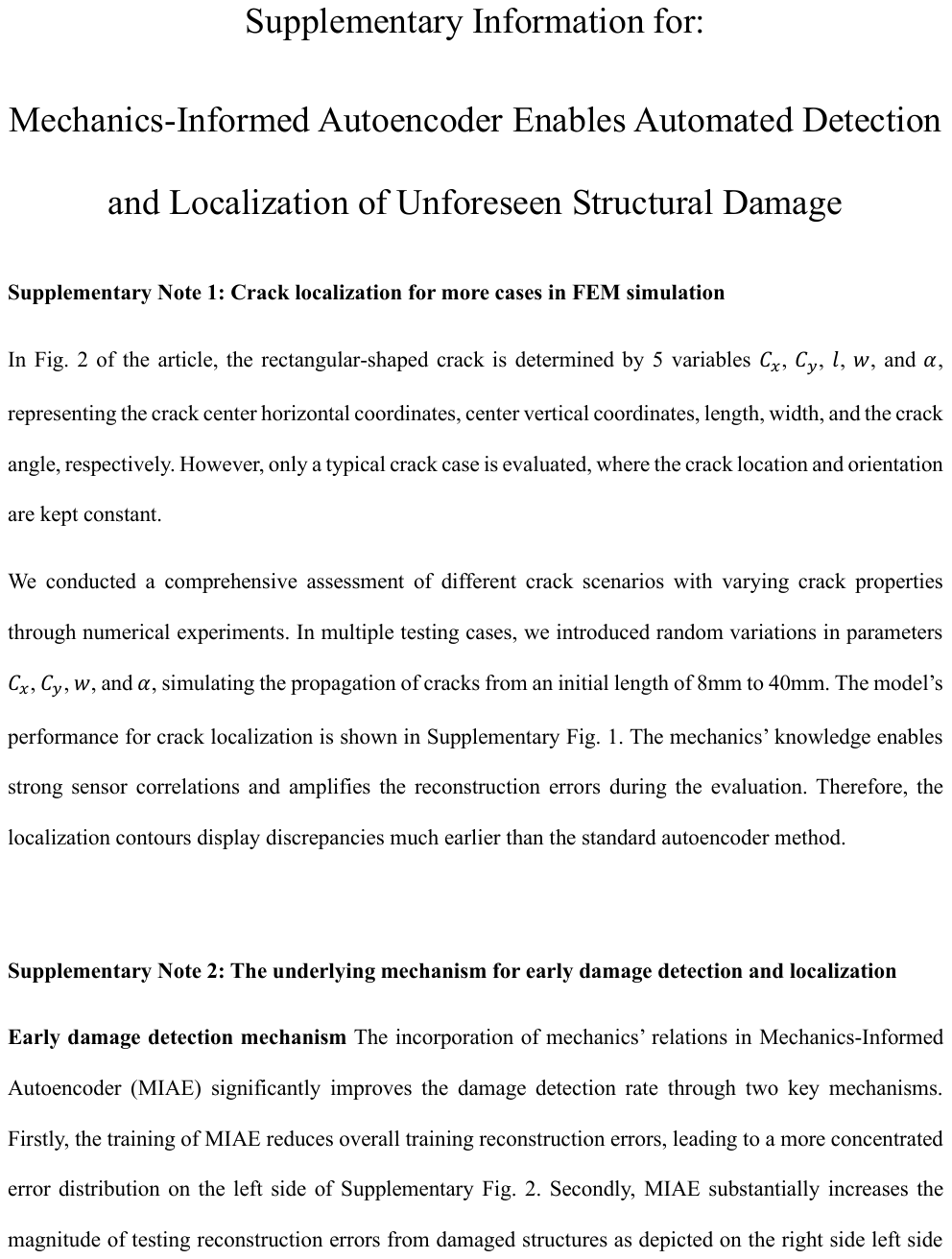} 
\includegraphics[page={2},scale=0.96,trim=2cm 0cm 0 2cm,clip]{Supplementary.pdf} 
\includegraphics[page={3},scale=0.96,trim=2cm 0cm 0 2cm,clip]{Supplementary.pdf} 
\includegraphics[page={4},scale=0.96,trim=2cm 0cm 0 2cm,clip]{Supplementary.pdf} 
\includegraphics[page={5},scale=0.96,trim=2cm 0cm 0 2cm,clip]{Supplementary.pdf} 
\includegraphics[page={6},scale=0.96,trim=2cm 0cm 0 2cm,clip]{Supplementary.pdf} 
\includegraphics[page={7},scale=0.96,trim=2cm 0cm 0 2cm,clip]{Supplementary.pdf} 
\includegraphics[page={8},scale=0.96,trim=2cm 0cm 0 2cm,clip]{Supplementary.pdf} 
\end{document}